\documentclass[11pt,a4paper]{article}
\usepackage[hyperref]{emnlp2018}
\usepackage{times}
\usepackage{latexsym}

\usepackage{url}

\aclfinalcopy 

\usepackage[utf8]{inputenc} 
\usepackage[T1]{fontenc}    
\usepackage{booktabs}       
\usepackage{amsfonts}       
\usepackage{nicefrac}       
\usepackage{microtype}      

\usepackage{amssymb,amsmath,amsthm,amsfonts}
\newtheorem{theorem}{Theorem}
\newtheorem{lemma}{Lemma}

\usepackage{graphicx}
\usepackage{amsmath}
\usepackage{subcaption}
\usepackage{algorithm}
\usepackage{algcompatible}
\usepackage{xcolor}
\usepackage[bbgreekl]{mathbbol}

\newcommand{\R}{\mathbb{R}}
\newcommand{\W}{\mathcal{W}}
\newcommand{\X}{\mathcal{X}}
\newcommand{\V}{\mathcal{V}}
\newcommand{\1}{\boldsymbol{1}}
\newcommand{\bv}{\boldsymbol{v}}
\newcommand{\bu}{\boldsymbol{u}}

\newcommand{\f}{\boldsymbol{f}}
\newcommand{\WMD}{\text{WMD}}
\newcommand{\softmin}{\mathit{softmin}}
\newcommand{\softmax}{\mathit{softmax}}
\newcommand{\T}{\mathsf{T}}
\newcommand{\E}{\mathcal{E}}

\title{Word Mover's Embedding: \\From Word2Vec to Document Embedding}

\author{
  Lingfei Wu \\
  IBM Research\\
  \texttt{wuli@us.ibm.com} \\
  \And
  Ian En-Hsu Yen  \\
  Carnegie Mellon University \\
  \texttt{eyan@cs.cmu.edu} \\
  \And
  Kun Xu \\
  IBM Research\\
  \texttt{kun.xu1@ibm.com} \\
  \AND
  Fangli Xu \\
  College of William and Mary \\
  \texttt{fxu02@email.wm.edu} \\
  \And
  Avinash Balakrishnan \\
  IBM Research\\
  \texttt{avinash.bala@us.ibm.com} \\
  \And
  Pin-Yu Chen \\
  IBM Research \\
  \texttt{Pin-yu.chen@ibm.com} \\
  \AND
  Pradeep Ravikumar \\
  Carnegie Mellon University \\
  \texttt{pradeepr@cs.cmu.edu} \\
  \And
  Michael J. Witbrock \\
  IBM Research\\
  \texttt{witbrock@us.ibm.com} \\
}

\date{}

\begin{document}
\maketitle

\begin{abstract}
While the celebrated Word2Vec technique yields semantically rich representations for individual words, there has been relatively less success in extending to generate unsupervised sentences or documents embeddings. Recent work has demonstrated that a distance measure between documents called \emph{Word Mover's Distance} (WMD) that aligns semantically similar words, yields unprecedented KNN classification accuracy. However, WMD is expensive to compute, and it is hard to extend its use beyond a KNN classifier. In this paper, we propose the \emph{Word Mover's Embedding } (WME), a novel approach to building an unsupervised document (sentence) embedding from pre-trained word embeddings. 
In our experiments on 9 benchmark text classification datasets and 22 textual similarity tasks, the proposed technique consistently matches or outperforms state-of-the-art techniques, with significantly higher accuracy on problems of short length.

\end{abstract}

\section{Introduction} 

Text representation plays an important role in many NLP-based tasks such as document classification and clustering \cite{zhang2018sentence,gui2016representative,gui2014estimate}, sense disambiguation \cite{gong2017prepositions,gong2018embedding}, machine translation \cite{mikolov2013exploiting}, document matching \cite{pham2015learning}, and sequential alignment \cite{peng2016recurrent,peng2015piefa}. Since there are no explicit features in text, much work has aimed to develop effective text representations. Among them, the simplest bag of words (BOW) approach \cite{salton1988term} and its term frequency variants (e.g. TF-IDF) \cite{robertson1994some} are most widely used due to simplicity, efficiency and often surprisingly high accuracy \cite{wang2012baselines}. However, simply treating words and phrases as discrete symbols fails to take into account word order and the semantics of the words, and suffers from frequent near-orthogonality due to its high dimensional sparse representation. To overcome these limitations, Latent Semantic Indexing  \cite{deerwester1990indexing} and Latent Dirichlet Allocation \cite{blei2003latent} were developed to extract more meaningful representations through singular value decomposition \cite{wu2015preconditioned} and learning a probabilistic BOW representation. 

A recent empirically successful body of research makes use of distributional or contextual information together with simple neural-network models to obtain vector-space representations of words and phrases \cite{bengio2003neural,mikolov2013efficient,mikolov2013distributed,pennington2014glove}. A number of researchers have proposed extensions of these towards learning semantic vector-space representations of sentences or documents. A simple but often effective approach is to use a weighted average over some or all of the embeddings of words in the document. While this is simple, important information could easily be lost in such a document representation, in part since it does not consider word order. A more sophisticated approach \cite{le2014distributed,Chen2017efficient} has focused on jointly learning embeddings for both words and paragraphs using  models similar to Word2Vec. However, these only use word order within a small context window; moreover, the quality of word embeddings learned in such a model may be limited by the size of the training corpus, which cannot scale to the large sizes used in the simpler word embedding models, and which may consequently weaken the quality of the document embeddings. 

Recently, Kusner et al. \cite{kusner2015word} presented a novel document distance metric, Word Mover's Distance (WMD), that measures the dissimilarity between two text documents in the Word2Vec embedding space.
Despite its state-of-the-art KNN-based classification accuracy over other methods, combining KNN and WMD incurs very high computational cost. More importantly, WMD is simply a distance that can be only combined with KNN or K-means, whereas many machine learning algorithms require a fixed-length feature representation as input.  

A recent work in building kernels from distance measures, D2KE (distances to kernels and embeddings)~\cite{wu2018d2ke} proposes a general methodology of the derivation of a positive-definite kernel from a given distance function, which enjoys better theoretical guarantees than other distance-based methods, such as $k$-nearest neighbor and distance substitution kernel \cite{haasdonk2004learning}, and has also been demonstrated to have strong empirical performance in the time-series domain \cite{wu2018random}. 

In this paper, we build on this recent innovation D2KE ~\cite{wu2018d2ke}, and present the \emph{Word Mover's Embedding} (WME), an unsupervised generic framework that learns continuous vector representations for text of variable lengths such as a sentence, paragraph, or document. In particular, we propose a new approach to first construct a positive-definite \emph{Word Mover's Kernel} via an infinite-dimensional feature map given by the Word Mover's distance (WMD) to random documents from a given distribution. Due to its use of the WMD, the feature map takes into account alignments of individual words between the documents in the semantic space given by the pre-trained word embeddings. Based on this kernel, we can then derive a document embedding via a Random Features approximation of the kernel, whose inner products approximate exact kernel computations. Our technique extends the theory of \emph{Random Features} to show convergence of the inner product between WMEs to a positive-definite kernel that can be interpreted as a soft version of (inverse) WMD. 

The proposed embedding is more efficient and flexible than WMD in many situations. As an example, WME with a simple linear classifier reduces the computational cost of WMD-based KNN \emph{from cubic to linear} in document length and \emph{from quadratic to linear} in number of samples, while simultaneously improving accuracy.
WME is extremely easy to implement, fully parallelizable, and highly extensible, since its two building blocks, Word2Vec and WMD, can be replaced by other techniques such as GloVe \cite{pennington2014glove,wieting2015ppdb} or S-WMD \cite{huang2016supervised}.  We evaluate WME on 9 real-world text classification tasks and 22 textual similarity tasks, and demonstrate that it consistently matches or outperforms other state-of-the-art techniques. Moreover, WME often achieves orders of magnitude speed-up compared to KNN-WMD while obtaining the same testing accuracy. Our code and data is available at {\small \url{https://github.com/IBM/WordMoversEmbeddings}}.


\begin{figure*}[htb]
\centering
	\begin{subfigure}[b]{0.45\textwidth}
      \includegraphics[width=\textwidth]{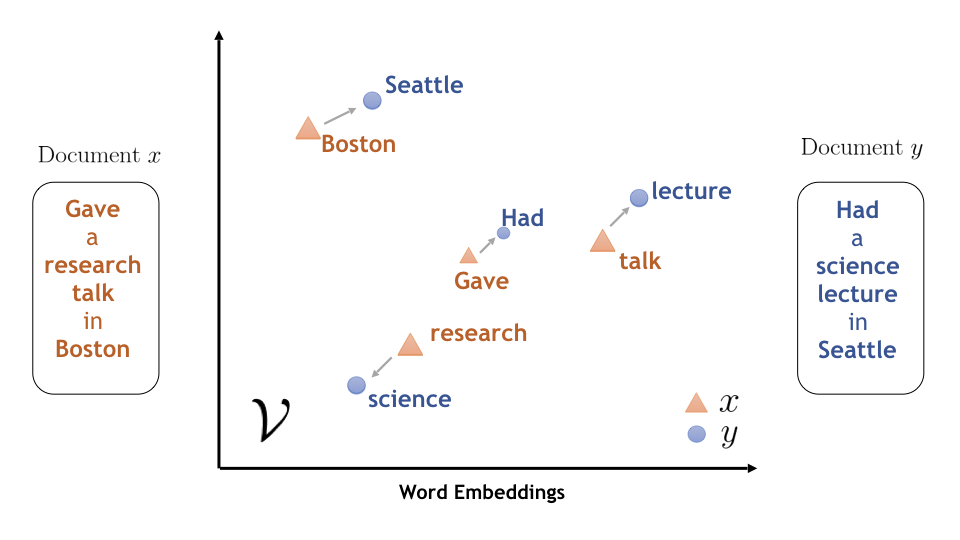}
      \caption{WMD}
      \label{fig:wmd}
    \end{subfigure}
	\begin{subfigure}[b]{0.45\textwidth}
      \includegraphics[width=\textwidth]{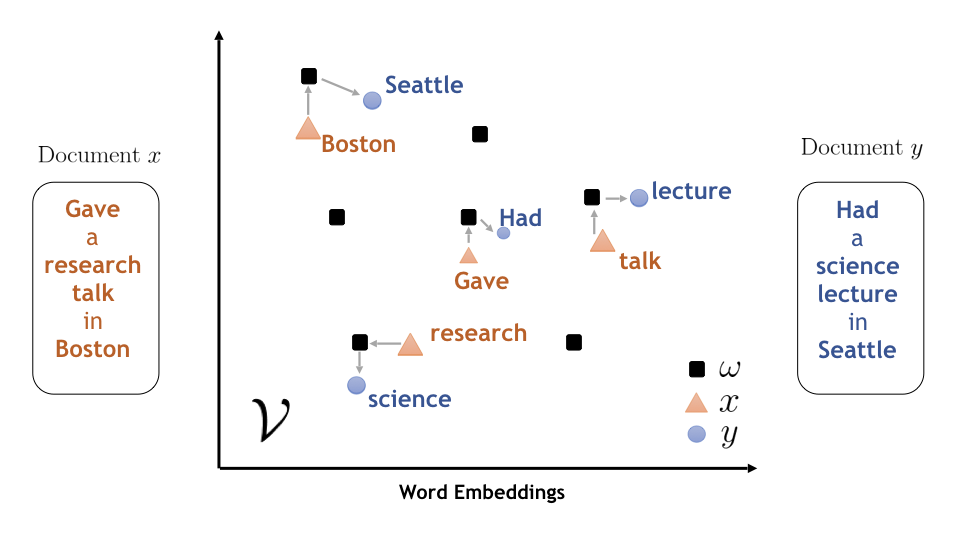}
      \caption{WME}
      \label{fig:wme}
    \end{subfigure}
\vspace{0mm}
\caption{An illustration of the WMD and WME. All non-stop words are marked as bold face. WMD measures the distance between two documents. WME approximates a kernel derived from WMD with a set of random documents.}
\label{fig:wmd_wme_demo}
\end{figure*}

\section{Word2Vec and Word Mover's Distance}
\label{sec:Word2Vec and WMD}

We briefly introduce Word2Vec and WMD, which are the key  building blocks of our proposed method. Here are some notations we will use throughout the paper. Given a total number of documents $N$ with a vocabulary $\W$ of size $|\W|=n$, the Word2vec embedding gives us a $d$-dimensional vector space $\V \subseteq \R^d$ such that any word in the vocabulary set $w\in \W$ is associated with a semantically rich vector representation $\bv_w\in \R^d$. Then in this work, we consider each document as a collection of word vectors $x:=(\bv_{j})_{j=1}^L$ and denote $\X:=\bigcup_{L=1}^{L_{\max}}\V^{L}$ as the space of documents.

\vskip0.1in
\noindent
\textbf{Word2Vec.} 
In the celebrated Word2Vec approach \cite{mikolov2013efficient,mikolov2013distributed}, two shallow yet effective models are used to learn vector-space representations of words (and phrases), by mapping those that co-occur frequently, and consequently with plausibly similar meaning, to nearby vectors in the embedding vector space. Due to the model's simplicity and scalability, high-quality word embeddings can be generated to capture a large number of precise syntactic and semantic word relationships by training over hundreds of billions of words and millions of named entities. The advantage of document representations building on top of word-level embeddings is that one can make full use of high-quality pre-trained word embeddings.
Throughout this paper we use Word2Vec as our first building block but other (unsupervised or supervised) word embeddings \cite{pennington2014glove,wieting2015ppdb} could also be utilized.

\vskip0.1in
\noindent
\textbf{Word Mover's Distance.} 
Word Mover's Distance was introduced by \cite{kusner2015word} as a special case of the Earth Mover's Distance \cite{rubner2000earth}, which can be computed as a solution of the well-known transportation problem \cite{hitchcock1941distribution,altschuler2017near}. WMD is a distance between two text documents $x$, $y\in\X$ that takes into account the alignments between words. Let $|x|$, $|y|$ be the number of distinct words in $x$ and $y$. Let $\f_x\in\R^{|x|}$, $\f_y\in\R^{|y|}$ denote the normalized frequency vectors of each word in the documents $x$ and $y$ respectively (so that  $\f_x^{\T}\1=\f_y^{\T}\1=1$). Then the WMD distance between documents $x$ and $y$ is defined as:
\begin{equation}\label{WMD}
\begin{aligned}
& \WMD(x,y) := \min_{F\in\R_+^{|x|\times |y|}} \langle C, F \rangle, \\
&s.t., F\1=\f_x,\;\;F^{\T}\1=\f_y.
\end{aligned}
\end{equation}
where $F$ is the transportation flow matrix with $F_{ij}$ denoting the amount of flow traveling from $i$-th word $x_i$ in $x$ to $j$-th word $y_j$ in $y$, and $C$ is the transportation cost with $C_{ij}:=\text{dist}(\bv_{x_i},\bv_{y_j})$ being the distance between two words measured in the Word2Vec embedding space. A popular choice is the Euclidean distance $\text{dist}(\bv_{x_i},\bv_{y_j})=\|\bv_{x_i}-\bv_{y_j}\|_2$. When $\text{dist}(\bv_{x_i},\bv_{y_j})$ is a \emph{metric}, the WMD distance in Eq. \eqref{WMD} also qualifies as a metric, and in particular, satisfies the triangle inequality~ \cite{rubner2000earth}. Building on top of Word2Vec, WMD is a particularly useful and accurate for measure of the distance between documents with semantically close but syntactically different words as illustrated in Figure 1(a). 

The WMD distance when coupled with KNN has been observed to have strong performance in classification tasks \cite{kusner2015word}. However, WMD is expensive to compute with computational complexity of $O(L^3 \log(L))$, especially for long documents where $L$ is large. Additionally, since WMD is just a document distance, rather than a document representation, using it within KNN incurs even higher computational costs $O(N^2L^3 \log(L))$. 

\section{Document Embedding via Word Mover's Kernel}

In this section, we extend the framework in \cite{wu2018d2ke}, to derive a positive-definite kernel from an alignment-aware document distance metric, which then gives us an unsupervised semantic embeddings of texts of variable length as a by-product through the theory of \emph{Random Feature Approximation} \cite{rahimi2007random}.

\subsection{Word Mover's Kernel}
We start by defining the \emph{Word Mover's Kernel}:
\begin{equation}\label{WMK}
\begin{aligned}
& k(x,y):=  \int p(\omega) \phi_{\omega}(x)\phi_{\omega}(y) d\omega, \\
& \text{where} \;\; \phi_{\omega}(x):=\exp(-\gamma \WMD(x,\omega)).
\end{aligned}
\end{equation}
where $\omega$ can be interpreted as a random document $\{\bv_j\}_{j=1}^D$ that contains a collection of random word vectors in $\V$, and $p(\omega)$ is a distribution over the space of all possible random documents $\Omega:=\bigcup_{D=1}^{D_{max}}\V^{D}$. $\phi_{\omega}(x)$ is an possibly infinite-dimensional feature map derived from the WMD between $x$ and all possible documents $\omega\in\Omega$.

An insightful interpretation of this kernel \eqref{WMK}:
\begin{equation*}\label{D2K2}
k(x,y):= \exp ( -\gamma\softmin_{p(\omega)} f(\omega) )
\end{equation*}
where
\begin{equation*}\label{softmin}
\softmin_{p(\omega)}\;f(\omega):= -\frac{1}{\gamma}\log \int p(\omega) e^{-\gamma f(\omega)} d\omega,
\end{equation*}
and 
$f(\omega) = \{ \WMD(x,\omega)\,+\,\WMD(\omega,y) \}$,
is a version of soft minimum function parameterized by $p(\omega)$ and $\gamma$. Comparing this with the usual definition of soft minimum
$
\softmin_i f_i := -\softmax(-f_i) = -\log \sum_{i} e^{-f_i},
$
it can be seen that the soft-min-variant in the above Equations uses a weighting of the objects $\omega$ via the probability density $p(\omega)$, and moreover has the additional parameter $\gamma$ to control the degree of smoothness. When $\gamma$ is large and $f(\omega)$ is Lipschitz-continuous, the value of the soft-min-variant is mostly determined by the minimum of $f(\omega)$. 

Note that since WMD is a metric, by the triangular inequality we have
\begin{equation*}\label{tri_ineq}
\WMD(x,y) \;\leq\;\min_{\omega\in \Omega}\; \left(\WMD(x,\omega)+\WMD(\omega,y)\right)
\end{equation*}
and the equality holds if we allow the length of random document $D_{\max}$ to be not smaller than $L$. Therefore, the kernel \eqref{WMK} serves as a good approximation to the WMD between any pair of documents $x$, $y$ as illustrated in Figure 1(b), while it is \emph{positive-definite} by the definition.

\subsection{Word Mover's Embedding}

Given the Word-Mover's Kernel in Eq. \eqref{WMK}, we can then use the Monte-Carlo approximation:
\begin{equation}\label{WME}
k(x,y) \approx \langle Z(x), Z(y) \rangle = \frac{1}{R}\sum_{i=1}^R \phi_{\omega_i}(x)\phi_{\omega_i}(y) 
\end{equation}
where $\{\omega_i\}_{i=1}^R$ are i.i.d. random documents drawn from $p(\omega)$ and $Z(x):=(\frac{1}{\sqrt{R}}\phi_{\omega_i}(x))_{i=1}^R$ gives a vector representation of document $x$. We call this random approximation \emph{Word Mover's Embedding}. Later, we show that this Random Features approximation in Eq. \eqref{WME} converges to the exact kernel \eqref{WMK} uniformly over all pairs of documents $(x,y)$ . 

\paragraph{Distribution $p(\omega)$.}
A key ingredient in the Word Mover's Kernel and Embedding is the distribution $p(\omega)$ over random documents. Note that $\omega \in \mathcal{X}$ consists of sets of words, each of which lies in the Word2Vec embedding space; the characteristics of which need to be captured by $p(\omega)$ in order to generate (sets of) ``meaningful'' random words. Several studies have found that the word vectors $\bv$ are roughly uniformly dispersed in the word embedding space~\cite{arora2016latent,arora2017simple}. This is also consistent with our empirical findings, that the uniform distribution centered by the mean of all word vectors in the documents is generally applicable for various text corpora. Thus, if $d$ is the dimensionality of the pre-trained word embedding space, we can draw a random word $u \in \mathbb{R}^d$ as $u_j \sim \text{Uniform}[v_{\min},v_{\max}]$, for $j = 1,\hdots,d$, and where $v_{\min}$ and $v_{\max}$ are some constants.

Given a distribution over random words, the remaining ingredient is the length $D$ of random documents. It is desirable to set these to a small number, in part because this length is indicative of the number of hidden global topics, and we expect the number of such global topics to be small. In particular, these global topics will allow short random documents to align with the documents to obtain ``topic-based'' discriminatory features. Since there is no prior information for global topics, we choose to uniformly sample the length of random documents as $D \sim \text{Uniform}[1, D_{\max}]$, for some constant $D_{\max}$. Stitching the distributions over words, and over the number of words, we then get a distribution over random documents. We note that our WME embedding allows potentially other random distributions, and other types of word embeddings, making it a flexible and powerful feature learning framework to utilize state-of-the-art techniques.  

\begin{algorithm}[tbh]
\caption{Word Mover's Embedding: An Unsupervised Feature Representation for Documents}
\label{alg:WME_features}
\begin{algorithmic}[1]
    \STATEx {\bf Input:} Texts $\{x_i\}_{i=1}^N$, $D_{\max}$, $R$.
    \STATEx {\bf Output:} Matrix $Z_{N \times R}$, with rows corresponding to text embeddings.
    \STATE Compute $v_{\max}$ and $v_{\min}$ as the maximum and minimum values, over all coordinates of the word vectors $\bv$ of $\{x_i\}_{i=1}^N$, from any pre-trained word embeddings (e.g. Word2Vec, GloVe or PSL999).
    \FOR {$j = 1, \ldots, R$}
        \STATE Draw $D_j \sim \text{Uniform}[1, D_{\max}]$. 
        \STATE Generate a random document $\omega_j$ consisting of $D_j$ number of random words drawn as $\omega_{j\ell}  \sim \text{Uniform}{[v_{\min},v_{\max}]}^d$, $\ell = 1,\hdots,D_j$.
        \STATE Compute $\f_{x_i}$ and $\f_{\omega_j}$ using a popular weighting scheme (e.g. NBOW or TF-IDF). 
        \STATE Compute the WME feature vector $Z_j = \phi_{\omega_j}(\{x_i\}_{i=1}^N)$ using WMD in Equation \eqref{WMK}.  
    \ENDFOR
    \STATE Return $Z(\{x_i\}_{i=1}^N) = \frac{1}{\sqrt{R}} [Z_1 \ Z_2 \ \ldots \ Z_R]$
\end{algorithmic}
\end{algorithm}

Algorithm \ref{alg:WME_features} summarizes the overall procedure to generate feature vectors for text of any length such as sentences, paragraphs, and documents.

\begin{table*}[ht]
\centering
\scriptsize
\caption{Properties of the datasets}
\label{tb:info of datasets}
\vspace{0mm}
\begin{center}
    \begin{tabular}{ c c c c c c c}
    \hline
    Dataset & $C$:Classes & $N$:Train & $M$:Test & BOW Dim & $L$:Length & Application \\ \hline 
    BBCSPORT   & 5 & 517 & 220 & 13243 & 117 & BBC sports article labeled by sport \\ 
    TWITTER	   & 3 & 2176 &	932 & 6344 & 9.9  & tweets categorized by sentiment \\ 
    RECIPE     & 15 & 3059 & 1311	& 5708 & 48.5 & recipe procedures labeled by origin \\ 
    OHSUMED    & 10 & 3999 & 5153 & 31789 & 59.2 & medical abstracts (class subsampled) \\ 
    CLASSIC    & 4 & 4965 & 2128 & 24277 & 38.6 & academic papers labeled by publisher \\ 
    REUTERS   & 8 & 5485 & 2189 & 22425 & 37.1 & news dataset (train/test split) \\ 
    AMAZON     & 4 & 5600 & 2400  & 42063 & 45.0 & amazon reviews labeled by product \\ 
    20NEWS    & 20 & 11293 & 7528 & 29671 & 72 & canonical user-written posts dataset \\ 
    RECIPE\_L & 20 & 27841 & 11933 & 3590 & 18.5 & recipe procedures labeled by origin \\ \hline
    \end{tabular}
\end{center}
\vspace{0mm}
\end{table*}

KNN-WMD, which uses the WMD distance together with KNN based classification, requires $O(N^2)$ evaluations of the WMD distance, which in turn has $O(L^3 \log(L))$ complexity, assuming that documents have lengths bounded by $L$, leading to an overall complexity of $O(N^2 \,L^3\, \log(L)$. In contrast, our WME approximation only requires super-linear complexity of $O(NRLlog(L))$ when $D$ is constant. This is because in our case each evaluation of WMD only requires $O(D^2 \, L \log(L))$ \cite{bourgeois1971extension}, due to the short length $D$ of our random documents. This dramatic reduction in computation significantly accelerates training and testing when combined with empirical risk minimization classifiers such as SVMs. A simple yet useful trick is to pre-compute the word distances to avoid redundant computations since a pair of words may appear multiple times in different pairs of documents. Note that the computation of the ground distance between each pair of word vectors in documents has a $O(L^2\,d)$ complexity, which could be close to one WMD evaluation if document length $L$ is short and word vector dimension $d$ is large. This simple scheme leads to additional improvement in runtime performance of our WME method that we show in our experiments.

\subsection{Convergence of WME}

In this section, we study the convergence of our embedding \eqref{WME} to the exact kernel \eqref{WMK} under the framework of Random Features (RF) approximation \cite{rahimi2007random}. Note that the standard RF convergence theory applies only to the shift-invariant kernel operated on two vectors, while our kernel \eqref{WMK} operates on two documents $x,y\in\X$ that are sets of word vectors. In \cite{wu2018d2ke}, a general RF convergence theory is provided for any \emph{distance-based kernel} as long as a finite covering number is given w.r.t. the given distance. In the following lemma, we provide the covering number for all documents of bounded length under the \emph{Word Mover's Distance}. Without loss of generality, we will assume that the word embeddings $\{\bv\}$ are normalized s.t. $\|\bv\|\leq 1$.

\begin{lemma} \label{lemma:convergence}
There exists an $\epsilon$-covering $\E$ of $\X$ under the WMD metric with Euclidean ground distance, so that:
$$
\forall x\in\X, \exists x_i\in \E,\;\WMD(x,x_i)\leq \epsilon,
$$
that has size bounded as $|\E| \leq (\frac{2}{\epsilon})^{d\,L}$, where $L$ is a bound on the length of document $x\in\X$.
\end{lemma}

Equipped with Lemma \ref{lemma:convergence}, we can derive the following convergence result as a simple corollary of the theoretical results in \cite{wu2018d2ke}. We defer the proof to the appendix \ref{App:Appendix A: Proof of Lemma and Theorem}.

\begin{theorem}\label{thm:convergence}
Let $\Delta_R(x,y)$ be the difference between the exact kernel \eqref{WMK} and the random approximation \eqref{WME} with $R$ samples, we have uniform convergence
\begin{equation*}\label{converge_result}
\begin{aligned}
P\left\{ \max_{x,y\in\X} |\Delta_R(x,y)| > 2t\right\} 
& \leq 2\left(\frac{12\gamma}{t}\right)^{2dL}e^{-Rt^2/2}.
\end{aligned}
\end{equation*}
where $d$ is the dimension of word embedding and $L$ is a bound on the document length. In other words, to guarantee $|\Delta_R(x,y)|\leq \epsilon$ with probability at least $1-\delta$, it suffices to have
$$
R = \Omega\biggl(\frac{dL}{\epsilon^2}\log(\frac{\gamma}{\epsilon})+\frac{1}{\epsilon^2}\log(\frac{1}{\delta}) \biggr).
$$
\end{theorem}

\section{Experiments}
\label{sec:Experiments}
We conduct an extensive set of experiments to demonstrate the effectiveness and efficiency of the proposed method. We first compare its performance against 7 unsupervised document embedding approaches over a wide range of text classification tasks, including sentiment analysis, news categorization, amazon review, and recipe identification. We use 9 different document corpora, with 8 of these drawn from \cite{kusner2015word,huang2016supervised}; Table \ref{tb:info of datasets} provides statistics of the different datasets. We further compare our method against 10 unsupervised, semi-supervised, and supervised document embedding approaches on the 22 datasets from SemEval semantic textual similarity tasks. Our code is implemented in Matlab, and we use C Mex for the computationally intensive components of WMD 
\cite{rubner2000earth}. 

\subsection{Effects of $R$ and $D$ on WME}
\label{sec:Effects of R and D on Random Features}

\noindent
\textbf{Setup.} We first perform experiments to investigate the behavior of the WME method by varying the number of Random Features $R$ and the length $D$ of random documents. The hyper-parameter $\gamma$ is set via cross validation on training set over the range $[0.01, \ 10]$. We simply fix the $D_{min}=1$, and vary $D_{max}$ over the range $[3, \ 21]$. Due to limited space, we only show selected subsets of our results, with the rest listed in the Appendix \ref{App:More results about effects of $R$ and $D$ on random documents}. 

\begin{figure}[ht]
\centering
		\begin{subfigure}[b]{0.23\textwidth}
      \includegraphics[width=\textwidth]{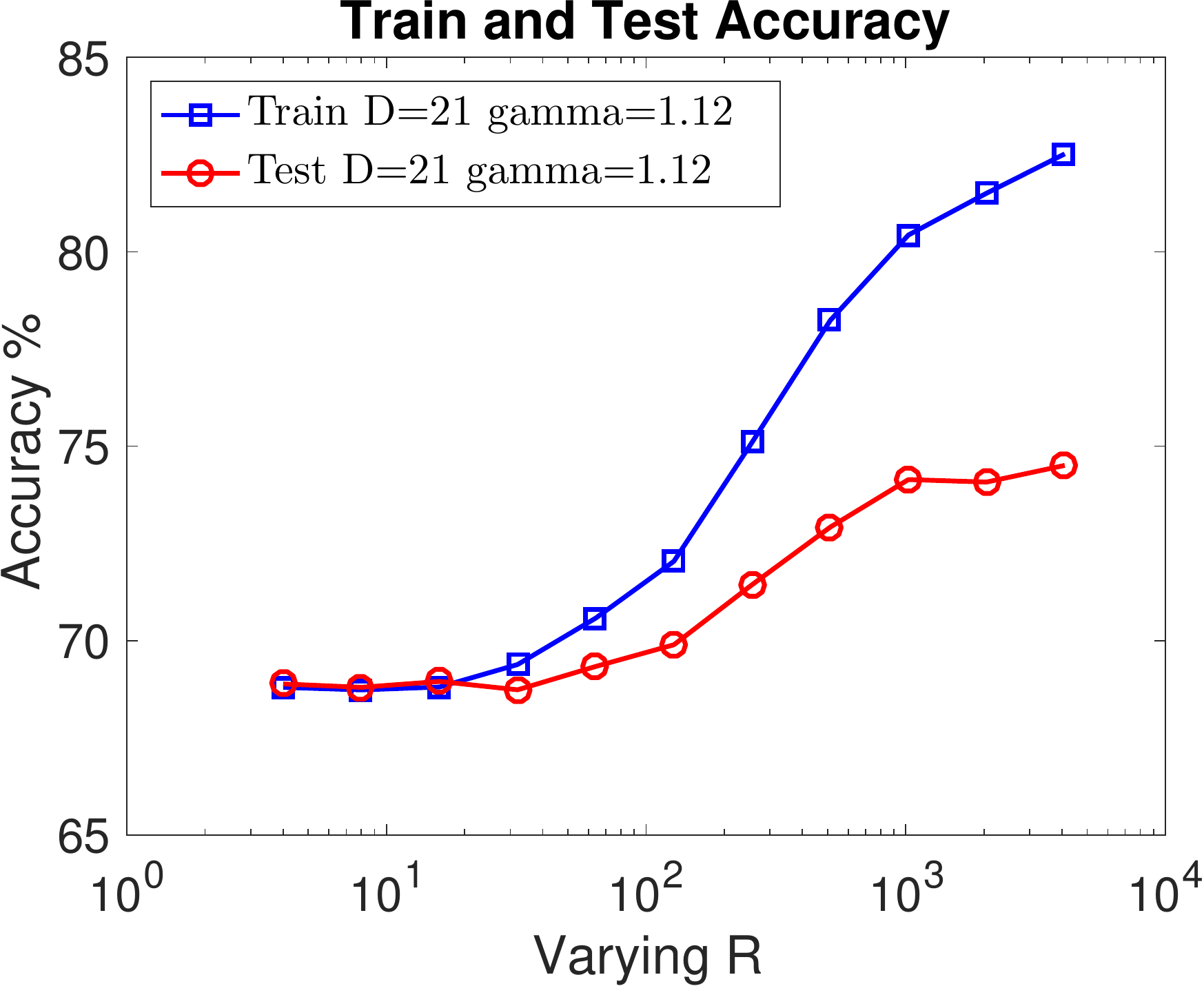}
      \caption{TWITTER}
      \label{fig:exptsA_varyingR_twitter}
      \end{subfigure}
      \begin{subfigure}[b]{0.23\textwidth}
      \includegraphics[width=\textwidth]{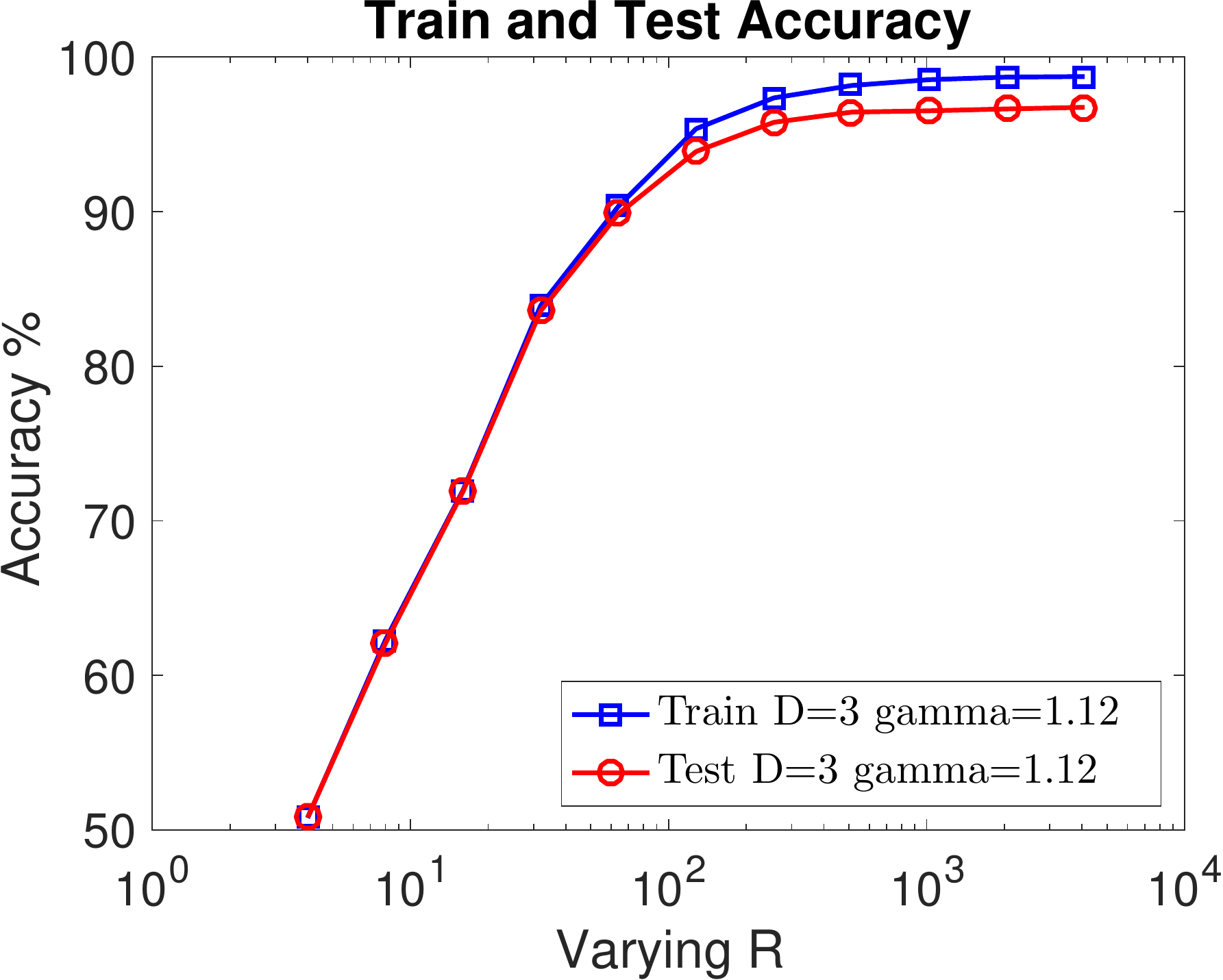}
      \caption{CLASSIC}
      \label{fig:exptsA_varyingR_classic}
     	\end{subfigure}
\vspace{0mm}
\caption{Train (Blue) and Test (Red) accuracy when varying $R$ with fixed $D$.}
\label{fig:exptsA_varyingR}
\vspace{0mm}
\end{figure}

\begin{figure}[ht]
\centering
		\begin{subfigure}[b]{0.23\textwidth}
      \includegraphics[width=\textwidth]{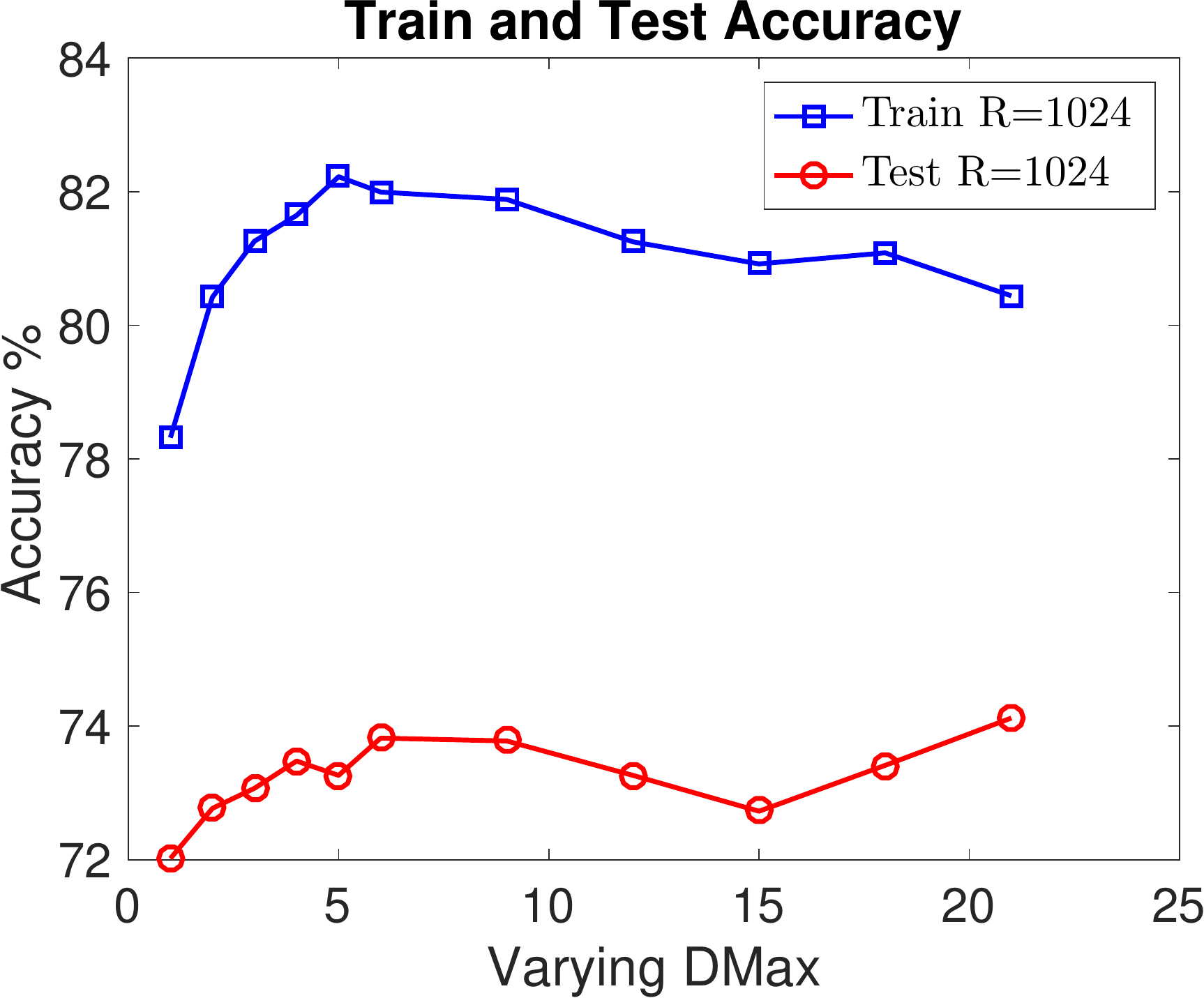}
      \caption{TWITTER}
      \label{fig:exptsA_varyingD_twitter}
      \end{subfigure}
      \begin{subfigure}[b]{0.23\textwidth}
      \includegraphics[width=\textwidth]{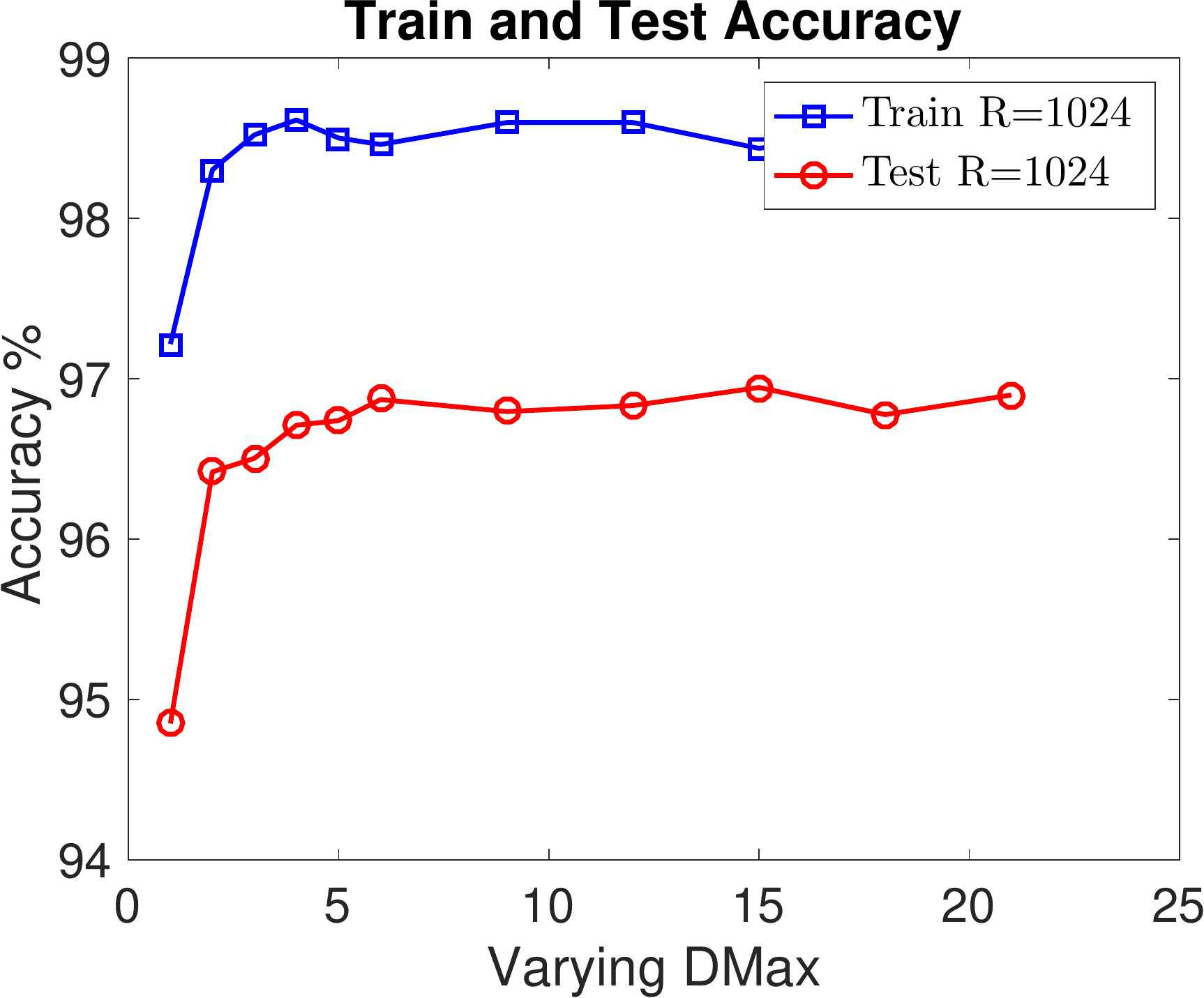}
      \caption{CLASSIC}
      \label{fig:exptsA_varyingD_classic}
     	\end{subfigure}
\vspace{0mm}
\caption{Train (Blue) and Test (Red) accuracy when varying $D$ with fixed $R$.}
\label{fig:exptsA_varyingD}
\vspace{0mm}
\end{figure}

\vskip0.05in
\noindent
\textbf{Effects of $R$.} We investigate how the performance changes when varying the number of Random Features $R$ from 4 to 4096 with fixed $D$. Fig.~\ref{fig:exptsA_varyingR} shows that both training and testing accuracies generally converge very fast when increasing $R$ from a small number ($R=4$) to a  relatively large number ($R=1024$), and then gradually reach to the optimal performance. This confirms our analysis in Theory \ref{thm:convergence} that the proposed WME  can guarantee the fast convergence to the exact kernel.

\vskip0.05in
\noindent
\textbf{Effects of $D$.} We further evaluate the training and testing accuracies when varying the length of random document $D$ with fixed $R$. As shown in Fig.~\ref{fig:exptsA_varyingD}, we can see that  near-peak performance can usually be achieved when $D$ is small (typically $D \leq 6$). This behavior illustrates two important aspects: (1) using very few random words (e.g. $D=1$) is not enough to generate useful Random Features when $R$ becomes large; (2) using too many random words (e.g. $D \geq 10$) tends to generate similar and redundant Random Features when increasing $R$. Conceptually, the number of random words in a random document can be thought of as the number of the global topics in documents, which is generally small. This is an important desired feature that confers both a performance boost as well as computational efficiency to the WME method.

\subsection{Comparison with KNN-WMD}
\label{sec:Comparisons against KNN-WMD in both accuracy and runtime}

\begin{table*}[ht]
\centering
\scriptsize
\caption{Test accuracy, and total training and testing time (in seconds) of WME against KNN-WMD. Speedups are computed between the best numbers of KNN-WMD+P and these of WME(SR)+P when achieving similar testing accuracy. Bold face highlights the best number for each dataset.}
\vspace{0mm}
\label{tb:comp_wme_knn_runtime}
\newcommand{\Bd}[1]{\textbf{#1}}
\begin{center}
    \begin{tabular}{ c ccc ccc ccc ccc c}
    \hline
    \multicolumn{1}{c}{Classifier}
    & \multicolumn{3}{c}{KNN-WMD \ \ KNN-WMD+P}
    & \multicolumn{3}{c}{WME(SR) \ \ WME(SR)+P}
    & \multicolumn{3}{c}{WME(LR) \ \ WME(LR)+P} 
    & \multicolumn{1}{c}{}\\ \hline 
    \multicolumn{1}{c}{Dataset}
	& Accu & Time & Time & Accu & Time & Time & Accu & Time & Time & Speedup\\ \hline
    BBCSPORT  & 95.4 $\pm$ 1.2  & 147 & 122 & 95.5 $\pm$ 0.7 & 3 & 1 & \Bd{98.2 $\pm$ 0.6} & 92 & 34 & \Bd{122}\\ 
    TWITTER  & 71.3 $\pm$ 0.6 & 25 & 4 & 72.5 $\pm$ 0.5 & 10 & 2 & \Bd{74.5 $\pm$ 0.5} & 162 & 34 & \Bd{2}\\ 
    RECIPE  & 57.4 $\pm$ 0.3 & 448 & 326 & 57.4 $\pm$ 0.5 & 18 & 4 & \Bd{61.8 $\pm$ 0.8} & 277 & 61 & \Bd{82} \\ 
    OHSUMED  & 55.5 & 3530 & 2807 & 55.8 & 24 & 7 & \Bd{64.5} & 757 & 240 & \Bd{401} \\ 
    CLASSIC  & \Bd{97.2 $\pm$ 0.1} & 777 & 520 & 96.6 $\pm$ 0.2 & 49 & 10 & 97.1 $\pm$ 0.4 & 388 & 70 & \Bd{52}\\  
    REUTERS  & 96.5 & 814 & 557 & 96.0 & 50 & 24 & \Bd{97.2} & 823 & 396 & \Bd{23}\\ 
    AMAZON  & 92.6 $\pm$ 0.3 & 2190 & 1319 & 92.7 $\pm$ 0.3 & 31 & 8 & \Bd{94.3 $\pm$ 0.4} & 495 & 123 & \Bd{165}\\ 
    20NEWS  & 73.2 & 37988 & 32610 & 72.9 & 205 & 69 & \Bd{78.3} & 1620 & 547 & \Bd{472} \\ 
    RECIPE\_L  & 71.4 $\pm$ 0.5 & 5942 & 2060 & 72.5 $\pm$ 0.4 & 113 & 20 & \Bd{79.2 $\pm$ 0.3} & 1838 & 330 & \Bd{103} \\ \hline
    \end{tabular}
\end{center}
\vspace{0mm}
\end{table*}

\begin{table*}[ht]
\centering
\scriptsize
\caption{Testing accuracy of WME against Word2Vec and Doc2Vec-based methods.}
\vspace{0mm}
\label{tb:comp_word2vec}
\newcommand{\Bd}[1]{\textbf{#1}}
\begin{center}
    \begin{tabular}{ c c c c c c c c}
    \hline
    Dataset & SIF(GloVe) & Word2Vec+nbow & Word2Vec+tf-idf & PV-DBOW & PV-DM & Doc2VecC & WME \\ \hline 
    BBCSPORT  & 97.3 $\pm$ 1.2 & 97.3 $\pm$ 0.9 & 96.9 $\pm$ 1.1 & 97.2 $\pm$ 0.7 & 97.9 $\pm$ 1.3  & 90.5 $\pm$ 1.7 & \Bd{98.2 $\pm$ 0.6} \\ 
    TWITTER  & 57.8  $\pm$ 2.5 & 72.0 $\pm$ 1.5 & 71.9 $\pm$ 0.7 & 67.8 $\pm$ 0.4 & 67.3 $\pm$ 0.3  & 71.0 $\pm$ 0.4 & \Bd{74.5 $\pm$ 0.5} \\ 
    OHSUMED & \Bd{67.1} & 63.0 & 60.6 & 55.9 & 59.8 & 63.4 & 64.5 \\ 
    CLASSIC & 92.7 $\pm$ 0.9 & 95.2 $\pm$ 0.4 & 93.9$\pm$ 0.4 & 97.0 $\pm$ 0.3 & 96.5 $\pm$ 0.7  & 96.6 $\pm$ 0.4  & \Bd{97.1 $\pm$ 0.4} \\
    REUTERS & 87.6 & 96.9 & 95.9 & 96.3 & 94.9  & 96.5 & \Bd{97.2} \\ 
    AMAZON & 94.1 $\pm$ 0.2 & 94.0 $\pm$ 0.5 & 92.2 $\pm$ 0.4 & 89.2 $\pm$ 0.3 & 88.6 $\pm$ 0.4  & 91.2 $\pm$ 0.5 & \Bd{94.3 $\pm$ 0.4} \\ 
    20NEWS & 72.3 & 71.7  & 70.2  & 71.0 & 74.0 & 78.2 & \Bd{78.3} \\ 
    RECIPE\_L  & 71.1 $\pm$ 0.5 & 74.9 $\pm$ 0.5 & 73.1 $\pm$ 0.6 & 73.1 $\pm$ 0.5 & 71.1 $\pm$ 0.4  & 76.1 $\pm$ 0.4  & \Bd{79.2 $\pm$ 0.3} \\ \hline
    \end{tabular}
\end{center}
\vspace{0mm}
\end{table*}

\vskip0.05in
\noindent
\textbf{Baselines.} We now compare two WMD-based methods in terms of testing accuracy and total training and testing runtime. 
We consider two variants of WME with different sizes of $R$. WME(LR) stands for WME with large rank that achieves the best accuracy (using $R$ up to 4096) with more computational time, while WME(SR) stands for WME with small rank that obtains comparable accuracy in less time. We also consider two variants of both methods where +P denotes that we precompute the ground distance between each pair of words to avoid redundant computations. 

\vskip0.05in
\noindent
\textbf{Setup.} Following \cite{kusner2015word,huang2016supervised}, for datasets that do not have a predefined train/test split, we report average and standard deviation of the testing accuracy and average run-time of the methods over five 70/30 train/test splits. For WMD, we provide the results (with respect to accuracy) from \cite{kusner2015word}; we also reran the experiments of KNN-WMD and found them to be  consistent with the reported results. For all methods, we perform 10-fold cross validation to search for the best parameters on the training documents. We employ a linear SVM implemented using LIBLINEAR \cite{fan2008liblinear} on WME since it can isolate the effectiveness of the feature representation from the power of the nonlinear learning solvers. For additional results on all KNN-based methods, please refer to Appendix~\ref{App:More results on Comparisons against distance-based methods}.

\vskip0.05in
\noindent
\textbf{Results.} Table \ref{tb:comp_wme_knn_runtime} corroborates the significant advantages of WME compared to KNN-WMD in terms of both accuracy and runtime. First, WME(SR) can consistently achieve better or similar accuracy compared to KNN-WMD while requiring order-of-magnitude less computational time on all datasets. Second, both methods can benefit from precomputation of the ground distance between a pair of words but WME gains much more from prefetch (typically 3-5x speedup). This is because the typical length $D$ of random documents is very short where computing ground distance between word vectors may be even more expensive than the corresponding WMD distance. Finally, WME(LR) can achieve much higher accuracy compared to KNN-WMD while still often requiring less computational time, especially on large datasets like 20NEWS and relatively long documents like OHSUMED. 

\begin{table*}[ht]
\centering
\scriptsize
\caption{Pearson's scores of WME against other unsupervised, semi-supervised, and supervised methods on 22 textual similarity tasks. Results are collected from \cite{arora2017simple} except our approach.}
\vspace{0mm}
\label{tb:comp_textual_similarity}
\newcommand{\Bd}[1]{\textbf{#1}}
\begin{center}
    \begin{tabular}{ c | cccccc | ccccc | cc}
    \hline
    \multicolumn{1}{c}{Approaches}  
    & \multicolumn{6}{c}{Supervised} 
    & \multicolumn{5}{c}{Unsupervised}
    & \multicolumn{2}{c}{Semi-supervised}\\ \hline 
    \multicolumn{1}{c}{WordEmbeddings}  
    & \multicolumn{6}{c}{PSL} 
    & \multicolumn{5}{c}{GloVe}
    & \multicolumn{2}{c}{PSL}\\ \hline 
    Tasks & PP & Dan & RNN & iRNN & LSTM(no) & LSTM(o.g.) & ST & nbow & tf-idf & SIF & WME & SIF & WME \\ \hline 
    STS'12 & 58.7 & 56.0 & 48.1 & 58.4 & 51.0 & 46.4 & 30.8 & 52.5 & 58.7 & 56.2 & 60.6 & 59.5 & \Bd{62.8}\\ 
    STS'13 & 55.8 & 54.2 & 44.7 & 56.7 & 45.2 & 41.5 & 24.8 & 42.3 & 52.1 & 56.6 & 54.5 & \Bd{61.8} & 56.3 \\ 
    STS'14 & 70.9 & 69.5 & 57.7 & 70.9 & 59.8 & 51.5 & 31.4 & 54.2 & 63.8 & 68.5 & 65.5 & \Bd{73.5} & 68.0 \\ 
    STS'15 & 75.8 & 72.7 & 57.2 & 75.6 & 63.9 & 56.0 & 31.0 & 52.7 & 60.6 & 71.7 & 61.8 & \Bd{76.3} & 64.2 \\
    SICK'14 & 71.6 & 70.7 & 61.2 & 71.2 & 63.9 & 59.0 & 49.8 & 65.9 & 69.4 & 72.2 & 68.0 & \Bd{72.9} & 68.1 \\
    Twitter'15 & 52.9 & \Bd{53.7} & 45.1 & 52.9 & 47.6 & 36.1 & 24.7 & 30.3 & 33.8 & 48.0 & 41.6 & 49.0 & 47.4 \\ \hline
    \end{tabular}
\end{center}
\vspace{0mm}
\end{table*}

\subsection{Comparisons with Word2Vec \& Doc2Vec}
\label{sec:Comparisons against Word2Vec and Doc2Vec-based document representations}

\noindent
\textbf{Baselines.} We compare against 6 document representations methods: 1) \emph{Smooth Inverse Frequency} (SIF) \cite{arora2017simple}: a recently proposed simple but tough to beat baseline for sentence embeddings, combining a new weighted scheme of word embeddings with dominant component removal; 2) \emph{Word2Vec+nbow}: a weighted average of word vectors using NBOW weights; 3) \emph{Word2Vec+tf-idf}: a weighted average of word vectors using TF-IDF weights; 4) \emph{PV-DBOW} \cite{le2014distributed}: distributed bag of words model of Paragraph Vectors; 5) \emph{PV-DM} \cite{le2014distributed}: distributed memory model of Paragraph Vectors; 6) \emph{Doc2VecC} \cite{Chen2017efficient}: a recently proposed document-embedding via corruptions,  achieving state-of-the-art performance in text classification. 

\vskip0.05in
\noindent
\textbf{Setup.} 
\emph{Word2Vec+nbow}, \emph{Word2Vec+tf-idf} and WME use pre-trained Word2Vec embeddings while SIF uses its default pre-trained GloVe embeddings. Following \cite{Chen2017efficient}, to enhance the performance of \emph{PV-DBOW}, \emph{PV-DM}, and \emph{Doc2VecC} these methods are trained transductively on both train and test, which is indeed beneficial for generating a better document representation (see Appendix \ref{App:More Results on Comparisons against Word2Vec and Doc2Vec-based document representations}). In contrast, the hyperparameters of WME are obtained through a 10-fold cross validation only on training set.
For a fair comparison, we run a linear SVM using LIBLINEAR on all methods.

\vskip0.05in
\noindent
\textbf{Results.} Table \ref{tb:comp_word2vec} shows that WME consistently outperforms or matches existing state-of-the-art document representation methods in terms of testing accuracy on all datasets except one (OHSUMED). The first highlight is that simple average of word embeddings often achieves better performance than \emph{SIF(Glove)}, indicating that removing the first principle component could hurt the expressive power of the resulting representation for some of classification tasks. Surprisingly, these two methods often achieve similar or better performance than \emph{PV-DBOW} and \emph{PV-DM}, which may be because of the high-quality pre-trained word embeddings. On the other hand, \emph{Doc2VecC} achieves much better testing accuracy than these previous methods on two datasets (20NEWS, and RECIPE\_L). This is mainly because that it benefits significantly from transductive training (See Appendix \ref{App:More Results on Comparisons against Word2Vec and Doc2Vec-based document representations}). Finally, the better performance of WME over these strong baselines stems from fact that WME is empowered by two important building blocks, WMD and Word2Vec, to yield a more informative representation of the documents by considering both the word alignments and the semantics of words. 

We refer the readers to additional results on the Imdb dataset in Appendix~\ref{App:More Results on Comparisons against Word2Vec and Doc2Vec-based document representations}, 
which also demonstrate the clear advantage of WME even compared to the supervised RNN method as well as the aforementioned baselines.

\subsection{Comparisons on textual similarity tasks}
\label{sec:Comparisons for performing textual similarity tasks}

\vskip0.05in
\noindent
\textbf{Baselines.} We compare WME against 10 supervised, simi-sepervised, and unsupervised methods for performing textual similarity tasks. 
Six supervised methods are initialized with Paragram-SL999(PSL) word vectors \cite{wieting2015ppdb} and then trained on the PPDB dataset, including: 1) \emph{PARAGRAM-PHRASE} (PP) \cite{wieting2015towards}: simple average of refined PSL word vectors; 2) \emph{Deep Averaging Network} (DAN) \cite{iyyer2015deep}; 3) \emph{RNN}: classical recurrent neural network; 4) \emph{iRNN}: a variant of RNN with the activation being the identify; 5) \emph{LSTM(no)} \cite{gers2002learning}: LSTM with no output gates; 6) \emph{LSTM(o.g.)} \cite{gers2002learning}: LSTM with output gates. 
Four unsupervised methods are: 1) \emph{Skip-Thought Vectors} (ST) \cite{kiros2015skip}: an encoder-decoder RNN model for generalizing Skip-gram to the sentence level; 2) \emph{nbow}: simple averaging of pre-trained GloVe word vectors; 3) \emph{tf-idf}: a weighted average of GloVe word vecors using TF-IDF weights; 4) \emph{SIF} \cite{arora2017simple}: a simple yet strong method on textual similarity tasks using GloVe word vecors.
Two semi-supervised methods use PSL word vectors, which are trained using labeled data \cite{wieting2015ppdb}.

\vskip0.05in
\noindent
\textbf{Setup.} There are total 22 textual similarity datasets from STS tasks (2012-2015) \cite{agirre2012semeval,agirre2013sem,agirre2014semeval,agirre2015semeval}, SemEval 2014 Semantic Relatedness task \cite{xu2015semeval}, and SemEval 2015 Twitter task \cite{marelli2014semeval}. 
The goal of these tasks is to predict the similarity between two input sentences. 
Each year STS usually has 4 to 6 different tasks and we only report the averaged Pearson's scores for clarity.
Detailed results on each dataset are listed in Appendix \ref{App:More results on comparisons for textual similarity tasks}. 

\vskip0.05in
\noindent
\textbf{Results.} Table \ref{tb:comp_textual_similarity} shows that WME consistently matches or outperforms other unsupervised and supervised methods except the \emph{SIF} method. Indeed, compared with \emph{ST} and \emph{nbow}, WME improves Pearson's scores substantially by 10\% to 33\% as a result of the consideration of word alignments and the use of TF-IDF weighting scheme. \emph{tf-idf} also improves over these two methods but is slightly worse than our method, indicating the importance of taking into account the alignments between the words. \emph{SIF} method is a strong baseline for textual similarity tasks but WME still can beat it on STS'12 and achieve close performance in other cases. Interestingly, WME is on a par with three supervised methods \emph{RNN}, \emph{LSTM(no)}, and \emph{LSTM(o.g.)} in most cases. The final remarks stem from the fact that, WME can gain significantly benefit from the supervised word embeddings similar to \emph{SIF}, both showing strong performance on PSL.


\section{Related Work}
\label{sec:Related Work}
Two broad classes of \emph{unsupervised} and \emph{supervised} methods have been proposed to generate sentence and document representations. The former primarily generate general purpose and domain independent embeddings of word sequences \cite{socher2011dynamic,kiros2015skip,arora2017simple}; many unsupervised training research efforts have focused on either training an auto-encoder to learn the latent structure of a sentence \cite{socher2013recursive}, a paragraph, or document \cite{li2015hierarchical}; or generalizing Word2Vec models to predict words in a paragraph \cite{le2014distributed,Chen2017efficient} or in neighboring sentences \cite{kiros2015skip}. 
However, some important information could be lost in the resulting document representation without considering the word order. Our proposed WME overcomes this difficulty by considering the alignments between each pair of words. 

The other line of work has focused on developing compositional supervised models to create a vector representation of sentences \cite{kim2016character,gong2018document}. Most of this work proposed composition using recursive neural networks based on parse structure \cite{socher2012semantic,socher2013recursive}, deep averaging networks over bag-of-words models \cite{iyyer2015deep,wieting2015towards}, convolutional neural networks \cite{kim2014convolutional,kalchbrenner2014convolutional,xu2018graph2seq}, and recurrent neural networks using long short-term memory \cite{tai2015improved,liu2015multi}. However, these methods are less well suited for domain adaptation settings.

\section{Conclusion}
 In this paper, we have proposed an alignment-aware text kernel using WMD for texts of variable lengths, which takes into account both word alignments and pre-trained high quality word embeddings in learning an effective semantics-preserving feature representation. The proposed WME is simple, efficient, flexible, and unsupervised. 
 Extensive experiments show that WME consistently matches or outperforms state-of-the-art models on various text classification and textual similarity tasks. WME embeddings can be easily used for a wide range of downstream supervised and unsupervised tasks. 

%

\bibliographystyle{acl_natbib_nourl}
\bibliography{WME_EMNLP18}

\begin{thebibliography}{62}
\expandafter\ifx\csname natexlab\endcsname\relax\def\natexlab#1{#1}\fi

\bibitem[{Agirre et~al.(2015)Agirre, Banea, Cardie, Cer, Diab, Gonzalez-Agirre,
  Guo, Lopez-Gazpio, Maritxalar, Mihalcea et~al.}]{agirre2015semeval}
Eneko Agirre, Carmen Banea, Claire Cardie, Daniel~M Cer, Mona~T Diab, Aitor
  Gonzalez-Agirre, Weiwei Guo, Inigo Lopez-Gazpio, Montse Maritxalar, Rada
  Mihalcea, et~al. 2015.
\newblock Semeval-2015 task 2: Semantic textual similarity, english, spanish
  and pilot on interpretability.
\newblock In \emph{SemEval@ NAACL-HLT}, pages 252--263.

\bibitem[{Agirre et~al.(2014)Agirre, Banea, Cardie, Cer, Diab, Gonzalez-Agirre,
  Guo, Mihalcea, Rigau, and Wiebe}]{agirre2014semeval}
Eneko Agirre, Carmen Banea, Claire Cardie, Daniel~M Cer, Mona~T Diab, Aitor
  Gonzalez-Agirre, Weiwei Guo, Rada Mihalcea, German Rigau, and Janyce Wiebe.
  2014.
\newblock Semeval-2014 task 10: Multilingual semantic textual similarity.
\newblock In \emph{SemEval@ COLING}, pages 81--91.

\bibitem[{Agirre et~al.(2013)Agirre, Cer, Diab, Gonzalez-Agirre, and
  Guo}]{agirre2013sem}
Eneko Agirre, Daniel Cer, Mona Diab, Aitor Gonzalez-Agirre, and Weiwei Guo.
  2013.
\newblock sem 2013 shared task: Semantic textual similarity, including a pilot
  on typed-similarity.
\newblock In \emph{In* SEM 2013: The Second Joint Conference on Lexical and
  Computational Semantics. Association for Computational Linguistics}.
  Citeseer.

\bibitem[{Agirre et~al.(2012)Agirre, Diab, Cer, and
  Gonzalez-Agirre}]{agirre2012semeval}
Eneko Agirre, Mona Diab, Daniel Cer, and Aitor Gonzalez-Agirre. 2012.
\newblock Semeval-2012 task 6: A pilot on semantic textual similarity.
\newblock In \emph{Proceedings of the First Joint Conference on Lexical and
  Computational Semantics-Volume 1: Proceedings of the main conference and the
  shared task, and Volume 2: Proceedings of the Sixth International Workshop on
  Semantic Evaluation}, pages 385--393. Association for Computational
  Linguistics.

\bibitem[{Altschuler et~al.(2017)Altschuler, Weed, and
  Rigollet}]{altschuler2017near}
Jason Altschuler, Jonathan Weed, and Philippe Rigollet. 2017.
\newblock Near-linear time approximation algorithms for optimal transport via
  sinkhorn iteration.
\newblock In \emph{Advances in Neural Information Processing Systems}, pages
  1964--1974.

\bibitem[{Arora et~al.(2016)Arora, Li, Liang, Ma, and
  Risteski}]{arora2016latent}
Sanjeev Arora, Yuanzhi Li, Yingyu Liang, Tengyu Ma, and Andrej Risteski. 2016.
\newblock A latent variable model approach to pmi-based word embeddings.
\newblock \emph{Transactions of the Association for Computational Linguistics},
  4:385--399.

\bibitem[{Arora et~al.(2017)Arora, Liang, and Ma}]{arora2017simple}
Sanjeev Arora, Yingyu Liang, and Tengyu Ma. 2017.
\newblock A simple but tough-to-beat baseline for sentence embeddings.
\newblock In \emph{ICLR}.

\bibitem[{Bengio et~al.(2003)Bengio, Ducharme, Vincent, and
  Jauvin}]{bengio2003neural}
Yoshua Bengio, R{\'e}jean Ducharme, Pascal Vincent, and Christian Jauvin. 2003.
\newblock A neural probabilistic language model.
\newblock \emph{Journal of machine learning research}, 3(Feb):1137--1155.

\bibitem[{Blei et~al.(2003)Blei, Ng, and Jordan}]{blei2003latent}
David~M Blei, Andrew~Y Ng, and Michael~I Jordan. 2003.
\newblock Latent dirichlet allocation.
\newblock \emph{Journal of machine Learning research}, 3(Jan):993--1022.

\bibitem[{Bourgeois and Lassalle(1971)}]{bourgeois1971extension}
Fran{\c{c}}ois Bourgeois and Jean-Claude Lassalle. 1971.
\newblock An extension of the munkres algorithm for the assignment problem to
  rectangular matrices.
\newblock \emph{Communications of the ACM}, 14(12):802--804.

\bibitem[{Buckley et~al.(1995)Buckley, Salton, Allan, and
  Singhal}]{buckley1995automatic}
Chris Buckley, Gerard Salton, James Allan, and Amit Singhal. 1995.
\newblock Automatic query expansion using smart: Trec 3.
\newblock \emph{NIST special publication sp}, pages 69--69.

\bibitem[{Chen(2017)}]{Chen2017efficient}
Minmin Chen. 2017.
\newblock Efficient vector representation for documents through corruption.
\newblock In \emph{ICLR}.

\bibitem[{Chen et~al.(2012)Chen, Xu, Weinberger, and
  Sha}]{chen2012marginalized}
Minmin Chen, Zhixiang Xu, Kilian Weinberger, and Fei Sha. 2012.
\newblock Marginalized denoising autoencoders for domain adaptation.
\newblock \emph{Proceedings of the 29th international conference on Machine
  learning}.

\bibitem[{Deerwester et~al.(1990)Deerwester, Dumais, Furnas, Landauer, and
  Harshman}]{deerwester1990indexing}
Scott Deerwester, Susan~T Dumais, George~W Furnas, Thomas~K Landauer, and
  Richard Harshman. 1990.
\newblock Indexing by latent semantic analysis.
\newblock \emph{Journal of the American society for information science},
  41(6):391.

\bibitem[{Fan et~al.(2008)Fan, Chang, Hsieh, Wang, and Lin}]{fan2008liblinear}
Rong-En Fan, Kai-Wei Chang, Cho-Jui Hsieh, Xiang-Rui Wang, and Chih-Jen Lin.
  2008.
\newblock Liblinear: A library for large linear classification.
\newblock \emph{Journal of machine learning research}, 9(Aug):1871--1874.

\bibitem[{Gers et~al.(2002)Gers, Schraudolph, and
  Schmidhuber}]{gers2002learning}
Felix~A Gers, Nicol~N Schraudolph, and J{\"u}rgen Schmidhuber. 2002.
\newblock Learning precise timing with lstm recurrent networks.
\newblock \emph{Journal of machine learning research}, 3(Aug):115--143.

\bibitem[{Glorot et~al.(2011)Glorot, Bordes, and Bengio}]{glorot2011domain}
Xavier Glorot, Antoine Bordes, and Yoshua Bengio. 2011.
\newblock Domain adaptation for large-scale sentiment classification: A deep
  learning approach.
\newblock In \emph{Proceedings of the 28th international conference on machine
  learning (ICML-11)}, pages 513--520.

\bibitem[{Gong et~al.(2018{\natexlab{a}})Gong, Bhat, and
  Viswanath}]{gong2018embedding}
Hongyu Gong, Suma Bhat, and Pramod Viswanath. 2018{\natexlab{a}}.
\newblock Embedding syntax and semantics of prepositions via tensor
  decomposition.
\newblock In \emph{Proceedings of the 2018 Conference of the North American
  Chapter of the Association for Computational Linguistics: Human Language
  Technologies, Volume 1 (Long Papers)}, volume~1, pages 896--906.

\bibitem[{Gong et~al.(2017)Gong, Mu, Bhat, and
  Viswanath}]{gong2017prepositions}
Hongyu Gong, Jiaqi Mu, Suma Bhat, and Pramod Viswanath. 2017.
\newblock Prepositions in context.
\newblock \emph{arXiv preprint arXiv:1702.01466}.

\bibitem[{Gong et~al.(2018{\natexlab{b}})Gong, Sakakini, Bhat, and
  Xiong}]{gong2018document}
Hongyu Gong, Tarek Sakakini, Suma Bhat, and JinJun Xiong. 2018{\natexlab{b}}.
\newblock Document similarity for texts of varying lengths via hidden topics.
\newblock In \emph{ACL}, volume~1, pages 2341--2351.

\bibitem[{Griffiths and Steyvers(2007)}]{griffiths2007probabilistic}
Tom Griffiths and Mark Steyvers. 2007.
\newblock Probabilistic topic models.
\newblock \emph{Latent Semantic Analysis: A Road to Meaning}.

\bibitem[{Gui et~al.(2016)Gui, Liu, Tao, Sun, and Tan}]{gui2016representative}
Jie Gui, Tongliang Liu, Dacheng Tao, Zhenan Sun, and Tieniu Tan. 2016.
\newblock Representative vector machines: a unified framework for classical
  classifiers.
\newblock \emph{IEEE transactions on cybernetics}, 46(8):1877--1888.

\bibitem[{Gui et~al.(2014)Gui, Sun, Cheng, Ji, and Wu}]{gui2014estimate}
Jie Gui, Zhenan Sun, Jun Cheng, Shuiwang Ji, and Xindong Wu. 2014.
\newblock How to estimate the regularization parameter for spectral regression
  discriminant analysis and its kernel version?
\newblock \emph{IEEE Transactions on Circuits and Systems for Video
  Technology}, 24(2):211--223.

\bibitem[{Haasdonk and Bahlmann(2004)}]{haasdonk2004learning}
Bernard Haasdonk and Claus Bahlmann. 2004.
\newblock Learning with distance substitution kernels.
\newblock In \emph{Joint Pattern Recognition Symposium}, pages 220--227.
  Springer.

\bibitem[{Hitchcock(1941)}]{hitchcock1941distribution}
Frank~L Hitchcock. 1941.
\newblock The distribution of a product from several sources to numerous
  localities.
\newblock \emph{Studies in Applied Mathematics}, 20(1-4):224--230.

\bibitem[{Huang et~al.(2016)Huang, Guo, Kusner, Sun, Sha, and
  Weinberger}]{huang2016supervised}
Gao Huang, Chuan Guo, Matt~J Kusner, Yu~Sun, Fei Sha, and Kilian~Q Weinberger.
  2016.
\newblock Supervised word mover's distance.
\newblock In \emph{Advances in Neural Information Processing Systems}, pages
  4862--4870.

\bibitem[{Iyyer et~al.(2015)Iyyer, Manjunatha, Boyd-Graber, and
  Daum{\'e}~III}]{iyyer2015deep}
Mohit Iyyer, Varun Manjunatha, Jordan Boyd-Graber, and Hal Daum{\'e}~III. 2015.
\newblock Deep unordered composition rivals syntactic methods for text
  classification.
\newblock In \emph{Proceedings of the 53rd Annual Meeting of the Association
  for Computational Linguistics and the 7th International Joint Conference on
  Natural Language Processing (Volume 1: Long Papers)}, volume~1, pages
  1681--1691.

\bibitem[{Kalchbrenner et~al.(2014)Kalchbrenner, Grefenstette, and
  Blunsom}]{kalchbrenner2014convolutional}
Nal Kalchbrenner, Edward Grefenstette, and Phil Blunsom. 2014.
\newblock A convolutional neural network for modelling sentences.
\newblock \emph{arXiv preprint arXiv:1404.2188}.

\bibitem[{Kim(2014)}]{kim2014convolutional}
Yoon Kim. 2014.
\newblock Convolutional neural networks for sentence classification.
\newblock \emph{arXiv preprint arXiv:1408.5882}.

\bibitem[{Kim et~al.(2016)Kim, Jernite, Sontag, and Rush}]{kim2016character}
Yoon Kim, Yacine Jernite, David Sontag, and Alexander~M Rush. 2016.
\newblock Character-aware neural language models.
\newblock In \emph{Thirtieth AAAI Conference on Artificial Intelligence}.

\bibitem[{Kiros et~al.(2015)Kiros, Zhu, Salakhutdinov, Zemel, Urtasun,
  Torralba, and Fidler}]{kiros2015skip}
Ryan Kiros, Yukun Zhu, Ruslan~R Salakhutdinov, Richard Zemel, Raquel Urtasun,
  Antonio Torralba, and Sanja Fidler. 2015.
\newblock Skip-thought vectors.
\newblock In \emph{Advances in neural information processing systems}, pages
  3294--3302.

\bibitem[{Kusner et~al.(2015)Kusner, Sun, Kolkin, and
  Weinberger}]{kusner2015word}
Matt Kusner, Yu~Sun, Nicholas Kolkin, and Kilian Weinberger. 2015.
\newblock From word embeddings to document distances.
\newblock In \emph{International Conference on Machine Learning}, pages
  957--966.

\bibitem[{Le and Mikolov(2014)}]{le2014distributed}
Quoc~V Le and Tomas Mikolov. 2014.
\newblock Distributed representations of sentences and documents.
\newblock In \emph{ICML}, volume~14, pages 1188--1196.

\bibitem[{Li et~al.(2015)Li, Luong, and Jurafsky}]{li2015hierarchical}
Jiwei Li, Minh-Thang Luong, and Dan Jurafsky. 2015.
\newblock A hierarchical neural autoencoder for paragraphs and documents.
\newblock \emph{arXiv preprint arXiv:1506.01057}.

\bibitem[{Liu et~al.(2015)Liu, Qiu, Chen, Wu, and Huang}]{liu2015multi}
Pengfei Liu, Xipeng Qiu, Xinchi Chen, Shiyu Wu, and Xuanjing Huang. 2015.
\newblock Multi-timescale long short-term memory neural network for modelling
  sentences and documents.
\newblock In \emph{Proceedings of the 2015 conference on empirical methods in
  natural language processing}, pages 2326--2335.

\bibitem[{Marelli et~al.(2014)Marelli, Bentivogli, Baroni, Bernardi, Menini,
  and Zamparelli}]{marelli2014semeval}
Marco Marelli, Luisa Bentivogli, Marco Baroni, Raffaella Bernardi, Stefano
  Menini, and Roberto Zamparelli. 2014.
\newblock Semeval-2014 task 1: Evaluation of compositional distributional
  semantic models on full sentences through semantic relatedness and textual
  entailment.
\newblock In \emph{SemEval@ COLING}, pages 1--8.

\bibitem[{Mikolov et~al.(2013{\natexlab{a}})Mikolov, Chen, Corrado, and
  Dean}]{mikolov2013efficient}
Tomas Mikolov, Kai Chen, Greg Corrado, and Jeffrey Dean. 2013{\natexlab{a}}.
\newblock Efficient estimation of word representations in vector space.
\newblock \emph{arXiv preprint arXiv:1301.3781}.

\bibitem[{Mikolov et~al.(2013{\natexlab{b}})Mikolov, Le, and
  Sutskever}]{mikolov2013exploiting}
Tomas Mikolov, Quoc~V Le, and Ilya Sutskever. 2013{\natexlab{b}}.
\newblock Exploiting similarities among languages for machine translation.
\newblock \emph{arXiv preprint arXiv:1309.4168}.

\bibitem[{Mikolov et~al.(2013{\natexlab{c}})Mikolov, Sutskever, Chen, Corrado,
  and Dean}]{mikolov2013distributed}
Tomas Mikolov, Ilya Sutskever, Kai Chen, Greg~S Corrado, and Jeff Dean.
  2013{\natexlab{c}}.
\newblock Distributed representations of words and phrases and their
  compositionality.
\newblock In \emph{Advances in neural information processing systems}, pages
  3111--3119.

\bibitem[{Peng et~al.(2016)Peng, Feris, Wang, and Metaxas}]{peng2016recurrent}
Xi~Peng, Rogerio~S Feris, Xiaoyu Wang, and Dimitris~N Metaxas. 2016.
\newblock A recurrent encoder-decoder network for sequential face alignment.
\newblock In \emph{European conference on computer vision}, pages 38--56.
  Springer, Cham.

\bibitem[{Peng et~al.(2015)Peng, Zhang, Yang, and Metaxas}]{peng2015piefa}
Xi~Peng, Shaoting Zhang, Yu~Yang, and Dimitris~N Metaxas. 2015.
\newblock Piefa: Personalized incremental and ensemble face alignment.
\newblock In \emph{Proceedings of the IEEE international conference on computer
  vision}, pages 3880--3888.

\bibitem[{Pennington et~al.(2014)Pennington, Socher, and
  Manning}]{pennington2014glove}
Jeffrey Pennington, Richard Socher, and Christopher~D Manning. 2014.
\newblock Glove: Global vectors for word representation.
\newblock In \emph{EMNLP}, volume~14, pages 1532--1543.

\bibitem[{Pham et~al.(2015)Pham, Luong, and Manning}]{pham2015learning}
Hieu Pham, Minh-Thang Luong, and Christopher~D Manning. 2015.
\newblock Learning distributed representations for multilingual text sequences.
\newblock In \emph{Proceedings of NAACL-HLT}, pages 88--94.

\bibitem[{Rahimi and Recht(2007)}]{rahimi2007random}
Ali Rahimi and Benjamin Recht. 2007.
\newblock Random features for large-scale kernel machines.
\newblock In \emph{Advances in Neural Information Processing Systems}, page~5.

\bibitem[{Robertson and Walker(1994)}]{robertson1994some}
Stephen~E Robertson and Steve Walker. 1994.
\newblock Some simple effective approximations to the 2-poisson model for
  probabilistic weighted retrieval.
\newblock In \emph{ACM SIGIR conference on Research and development in
  information retrieval}.

\bibitem[{Robertson et~al.(1995)Robertson, Walker, Jones, Hancock-Beaulieu,
  Gatford et~al.}]{robertson1995okapi}
Stephen~E Robertson, Steve Walker, Susan Jones, Micheline~M Hancock-Beaulieu,
  Mike Gatford, et~al. 1995.
\newblock Okapi at trec-3.
\newblock \emph{Nist Special Publication Sp}, 109:109.

\bibitem[{Rubner et~al.(2000)Rubner, Tomasi, and Guibas}]{rubner2000earth}
Yossi Rubner, Carlo Tomasi, and Leonidas~J Guibas. 2000.
\newblock The earth mover's distance as a metric for image retrieval.
\newblock \emph{International journal of computer vision}, 40(2):99--121.

\bibitem[{Salton and Buckley(1988)}]{salton1988term}
Gerard Salton and Christopher Buckley. 1988.
\newblock Term-weighting approaches in automatic text retrieval.
\newblock \emph{Information processing \& management}, 24(5):513--523.

\bibitem[{Socher et~al.(2011)Socher, Huang, Pennin, Manning, and
  Ng}]{socher2011dynamic}
Richard Socher, Eric~H Huang, Jeffrey Pennin, Christopher~D Manning, and
  Andrew~Y Ng. 2011.
\newblock Dynamic pooling and unfolding recursive autoencoders for paraphrase
  detection.
\newblock In \emph{Advances in neural information processing systems}, pages
  801--809.

\bibitem[{Socher et~al.(2012)Socher, Huval, Manning, and
  Ng}]{socher2012semantic}
Richard Socher, Brody Huval, Christopher~D Manning, and Andrew~Y Ng. 2012.
\newblock Semantic compositionality through recursive matrix-vector spaces.
\newblock In \emph{EMNLP}, pages 1201--1211. Association for Computational
  Linguistics.

\bibitem[{Socher et~al.(2013)Socher, Perelygin, Wu, Chuang, Manning, Ng, Potts
  et~al.}]{socher2013recursive}
Richard Socher, Alex Perelygin, Jean~Y Wu, Jason Chuang, Christopher~D Manning,
  Andrew~Y Ng, Christopher Potts, et~al. 2013.
\newblock Recursive deep models for semantic compositionality over a sentiment
  treebank.
\newblock In \emph{EMNLP}.

\bibitem[{Tai et~al.(2015)Tai, Socher, and Manning}]{tai2015improved}
Kai~Sheng Tai, Richard Socher, and Christopher~D Manning. 2015.
\newblock Improved semantic representations from tree-structured long
  short-term memory networks.
\newblock \emph{arXiv preprint arXiv:1503.00075}.

\bibitem[{Wang and Manning(2012)}]{wang2012baselines}
Sida Wang and Christopher~D Manning. 2012.
\newblock Baselines and bigrams: Simple, good sentiment and topic
  classification.
\newblock In \emph{Proceedings of the 50th Annual Meeting of the Association
  for Computational Linguistics: Short Papers-Volume 2}, pages 90--94.
  Association for Computational Linguistics.

\bibitem[{Wieting et~al.(2015{\natexlab{a}})Wieting, Bansal, Gimpel, and
  Livescu}]{wieting2015towards}
John Wieting, Mohit Bansal, Kevin Gimpel, and Karen Livescu.
  2015{\natexlab{a}}.
\newblock Towards universal paraphrastic sentence embeddings.
\newblock \emph{arXiv preprint arXiv:1511.08198}.

\bibitem[{Wieting et~al.(2015{\natexlab{b}})Wieting, Bansal, Gimpel, Livescu,
  and Roth}]{wieting2015ppdb}
John Wieting, Mohit Bansal, Kevin Gimpel, Karen Livescu, and Dan Roth.
  2015{\natexlab{b}}.
\newblock From paraphrase database to compositional paraphrase model and back.
\newblock \emph{Transactions of the ACL (TACL)}.

\bibitem[{Wu et~al.(2017)Wu, Romero, and Stathopoulos}]{wu2017primme_svds}
Lingfei Wu, Eloy Romero, and Andreas Stathopoulos. 2017.
\newblock Primme\_svds: A high-performance preconditioned svd solver for
  accurate large-scale computations.
\newblock \emph{SIAM Journal on Scientific Computing}, 39(5):S248--S271.

\bibitem[{Wu and Stathopoulos(2015)}]{wu2015preconditioned}
Lingfei Wu and Andreas Stathopoulos. 2015.
\newblock A preconditioned hybrid svd method for accurately computing singular
  triplets of large matrices.
\newblock \emph{SIAM Journal on Scientific Computing}, 37(5):S365--S388.

\bibitem[{Wu et~al.(2018{\natexlab{a}})Wu, Yen, Xu, Ravikumar, and
  Michael}]{wu2018d2ke}
Lingfei Wu, Ian En-Hsu Yen, Fnagli Xu, Pradeep Ravikumar, and Witbrock Michael.
  2018{\natexlab{a}}.
\newblock D2ke: From distance to kernel and embedding.
\newblock \emph{https://arxiv.org/abs/1802.04956}.

\bibitem[{Wu et~al.(2018{\natexlab{b}})Wu, Yen, Yi, Xu, Lei, and
  Witbrock}]{wu2018random}
Lingfei Wu, Ian En-Hsu Yen, Jinfeng Yi, Fangli Xu, Qi~Lei, and Michael
  Witbrock. 2018{\natexlab{b}}.
\newblock Random warping series: A random features method for time-series
  embedding.
\newblock In \emph{International Conference on Artificial Intelligence and
  Statistics}, pages 793--802.

\bibitem[{Xu et~al.(2018)Xu, Wu, Wang, and Sheinin}]{xu2018graph2seq}
Kun Xu, Lingfei Wu, Zhiguo Wang, and Vadim Sheinin. 2018.
\newblock Graph2seq: Graph to sequence learning with attention-based neural
  networks.
\newblock \emph{arXiv preprint arXiv:1804.00823}.

\bibitem[{Xu et~al.(2015)Xu, Callison-Burch, and Dolan}]{xu2015semeval}
Wei Xu, Chris Callison-Burch, and Bill Dolan. 2015.
\newblock Semeval-2015 task 1: Paraphrase and semantic similarity in twitter
  (pit).
\newblock In \emph{SemEval@ NAACL-HLT}, pages 1--11.

\bibitem[{Zhang et~al.(2018)Zhang, Liu, and Song}]{zhang2018sentence}
Yue Zhang, Qi~Liu, and Linfeng Song. 2018.
\newblock Sentence-state lstm for text representation.
\newblock \emph{arXiv preprint arXiv:1805.02474}.

\end{thebibliography}

\clearpage
\appendix
\section{Appendix A: Proof of Lemma \ref{lemma:convergence} and Theorem \ref{thm:convergence}}
\label{App:Appendix A: Proof of Lemma and Theorem}

\subsection{Proof of Lemma \ref{lemma:convergence}}
\label{App:Proof of Lemma 1}

\begin{proof}
Firstly, we find an $\epsilon$-covering $\E_W$ of size $(\frac{2}{\epsilon})^{d}$ for the word vector space $\V$ . Then define $\E$ as all possible sets of $\bv\in\E_W$ of size no larger than $L_{\max}$. We have $|\E|\leq (\frac{2}{\epsilon})^{dL_{\max}}$, and for any document $x=(\bv_j)_{j=1}^L \in\X$, we can find $x_i\in \E$ with also $L$ words $(\bu_j)_{j=1}^L$ such that $\|\bu_j-\bv_j\|\leq \epsilon$. Then by the definition of WMD \eqref{WMD}, a solution that assigns each word $\bv_j$ in $x$ to the word $\bu_j$ in $x_i$ would have overall cost less than $\epsilon$, and therefore, $\WMD(x,x_i)\leq \epsilon$.
\end{proof}

\subsection{Proof of Theorem \ref{thm:convergence}}
\label{App:Proof of Theorem 1}

\begin{proof}
Let $s_R(x,y)$ be the random approximation \eqref{WME}. Our goal is to bound the magnitude of $\Delta_R(x,y)=s_R(x,y)-k(x,y)$. Since $E[\Delta_R(x,y)]=0$ and $|\Delta_R(x,y)|\leq 1$, from Hoefding inequality, we have
$$
P\left\{ |\Delta_R(x,y)|\geq t \right\} \leq 2 \exp(-Rt^2/2)
$$
for a given pair of documents $(x,y)$. To get a uniform bound that holds for $\forall (x,y)\in\X\times\X$, we find an $\epsilon$-covering of $\X$ of finite size, given by Lemma \ref{lemma:convergence}. Applying union bound over the $\epsilon$-covering $\E$ for $x$ and $y$, we have
\begin{equation}\label{tmp1}
\begin{aligned}
&P\left\{ \max_{x_i\in\E,y_j\in\E} |\Delta_R(x_i,y_j)| > t \right\} \\
&\leq 2|\E|^2\exp(-Rt^2/2).
\end{aligned}
\end{equation}
Then by the definition of $\E$ we have $|\WMD(x,\omega)-\WMD(x_i,\omega)|\leq \WMD(x,x_i)\leq \epsilon$. Together with the fact that $\exp(-\gamma t)$ is Lipschitz-continuous with parameter $\gamma$ for $t\geq0$, we have 
$$
|\phi_{\omega}(x)-\phi_{\omega}(x_i)|\leq \gamma\epsilon
$$
and thus
$$
|s_R(x,y)-s_R(x_i,y_i)|\leq 3\gamma\epsilon, 
$$
and
$$
|k(x,y)-k(x_i,y_i)|\leq 3\gamma\epsilon
$$
for $\gamma\epsilon$ chosen to be $\leq 1$. This gives us 
\begin{equation}\label{tmp2}
|\Delta_R(x,y)-\Delta_R(x_i,y_i)|\leq 6\gamma\epsilon
\end{equation}
Combining \eqref{tmp1} and \eqref{tmp2}, we have
\begin{equation}\label{tmp3}
\begin{aligned}
& P\left\{ \max_{x_i\in\E,y_j\in\E} |\Delta_R(x,y)| > t + 6\gamma \epsilon\right\} \\ 
& \leq 2\left(\frac{2}{\epsilon}\right)^{2dL_{\max}}\exp(-Rt^2/2).
\end{aligned}
\end{equation}
Choosing $\epsilon=t/6\gamma$ yields the result.
\end{proof}

\section{Appendix B: Additional Experimental Results and Details}
\label{App:Appendix B: Additional Experimental Results and Details}

\begin{figure*}[htb]
\centering
		\begin{subfigure}[b]{0.23\textwidth}
      \includegraphics[width=\textwidth]{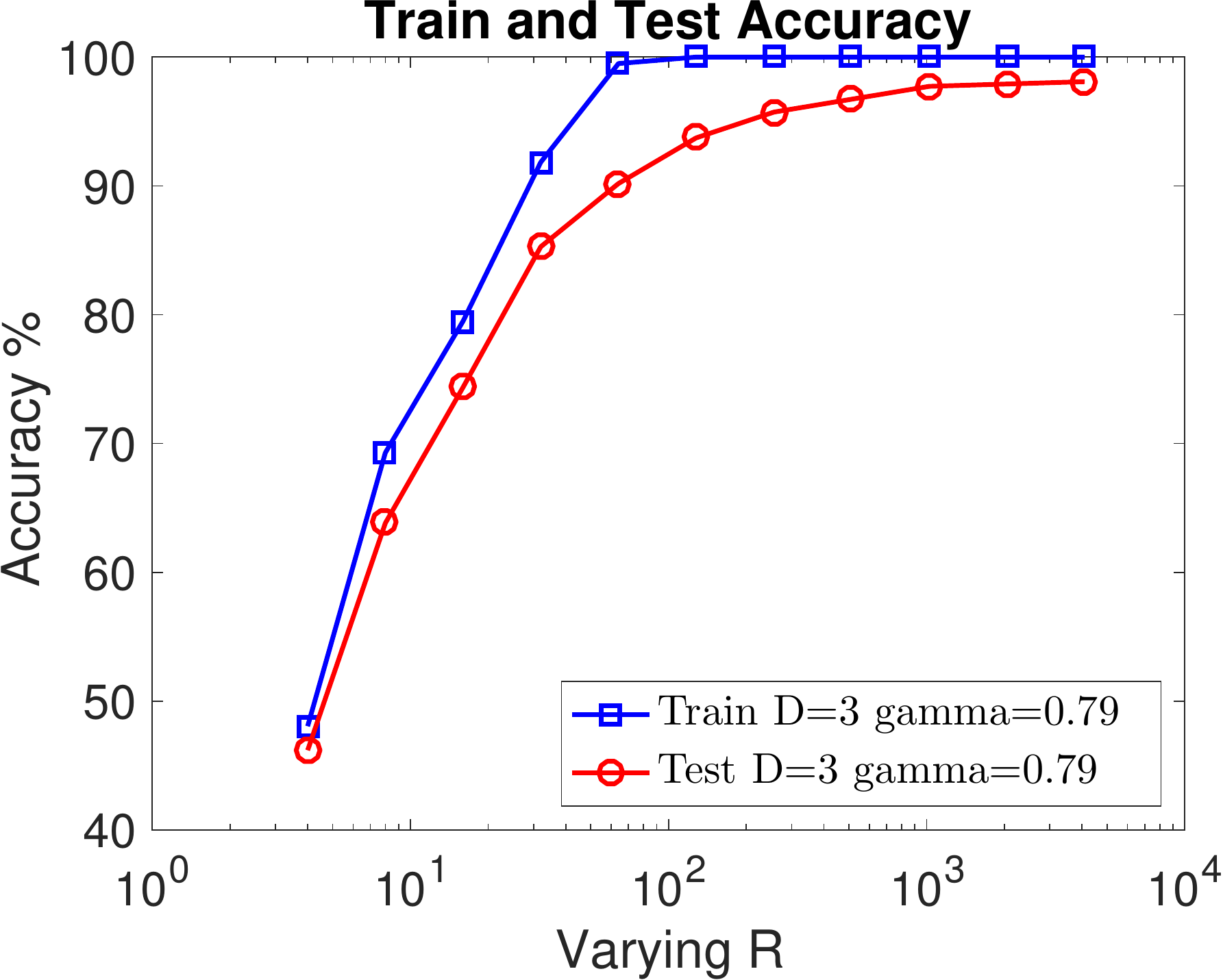}
      \caption{BBCSPORT}
      \label{App:fig:exptsA_varyingR_bbcsport}
     	\end{subfigure}
		\begin{subfigure}[b]{0.23\textwidth}
      \includegraphics[width=\textwidth]{Graphs/wmdk_varyingR/twitter_random_VaryingR_allSplits_CV_R512-eps-converted-to.pdf}
      \caption{TWITTER}
      \label{App:fig:exptsA_varyingR_twitter}
      \end{subfigure}
		\begin{subfigure}[b]{0.23\textwidth}
      \includegraphics[width=\textwidth]{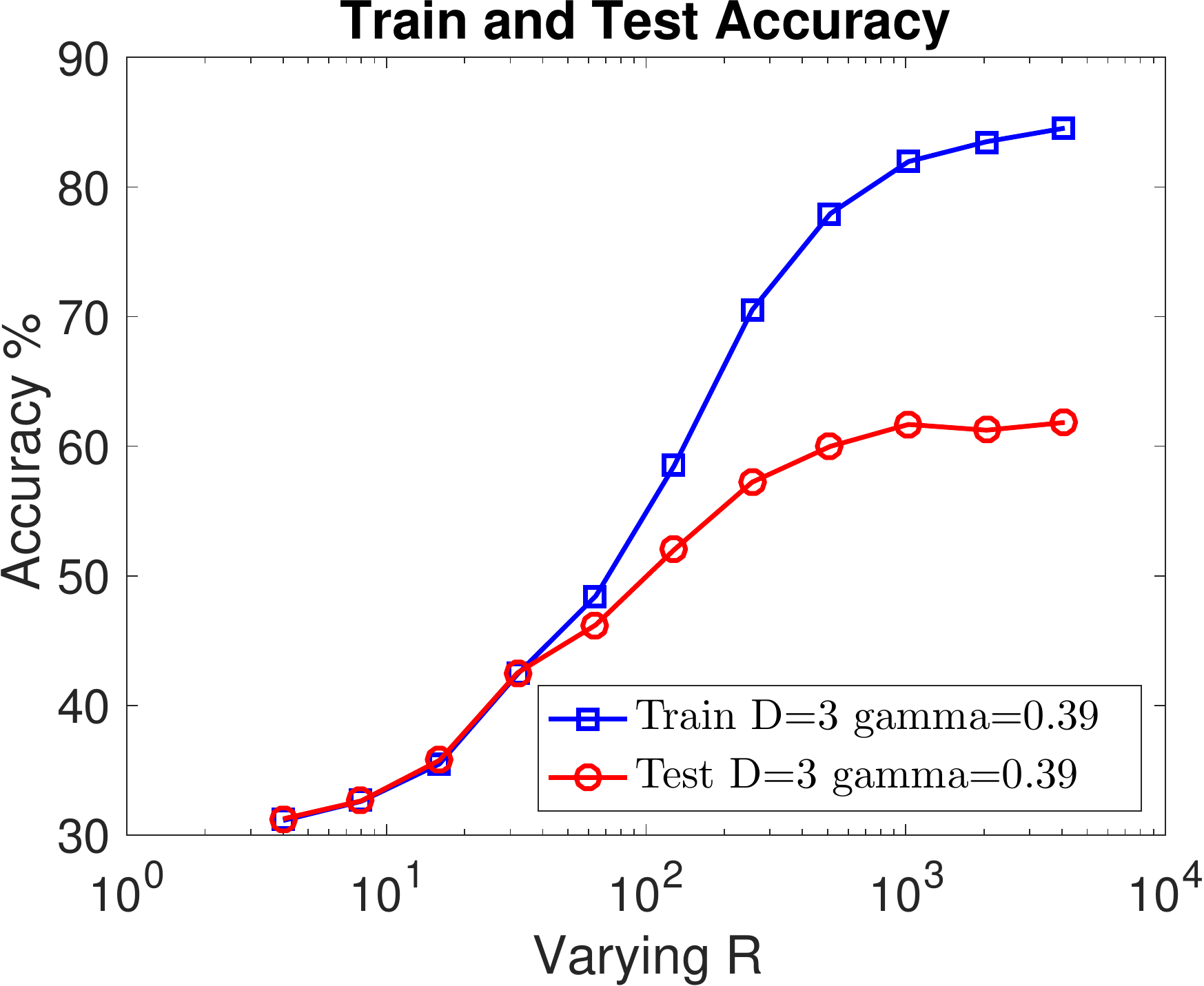}
      \caption{RECIPE}
      \label{App:fig:exptsA_varyingR_recipe2}
      \end{subfigure}
		\begin{subfigure}[b]{0.23\textwidth}
      \includegraphics[width=\textwidth]{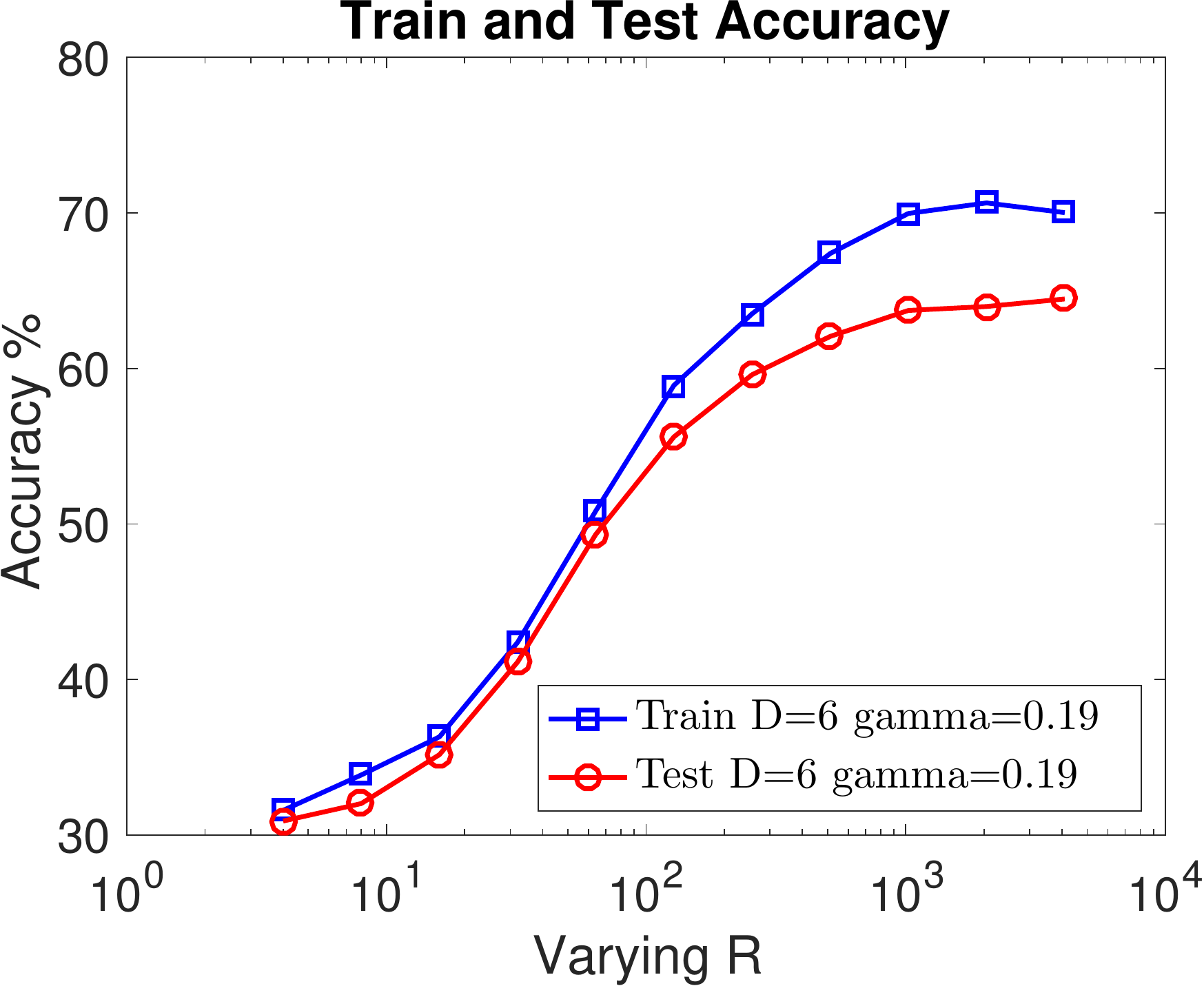}
      \caption{OHSUMED}
      \label{App:fig:exptsA_varyingR_ohsumed}
      \end{subfigure}
		\begin{subfigure}[b]{0.23\textwidth}
      \includegraphics[width=\textwidth]{Graphs/wmdk_varyingR/classic_random_VaryingR_allSplits_CV_R512-eps-converted-to.pdf}
      \caption{CLASSIC}
      \label{App:fig:exptsA_varyingR_classic}
      \end{subfigure}
      \begin{subfigure}[b]{0.23\textwidth}
      \includegraphics[width=\textwidth]{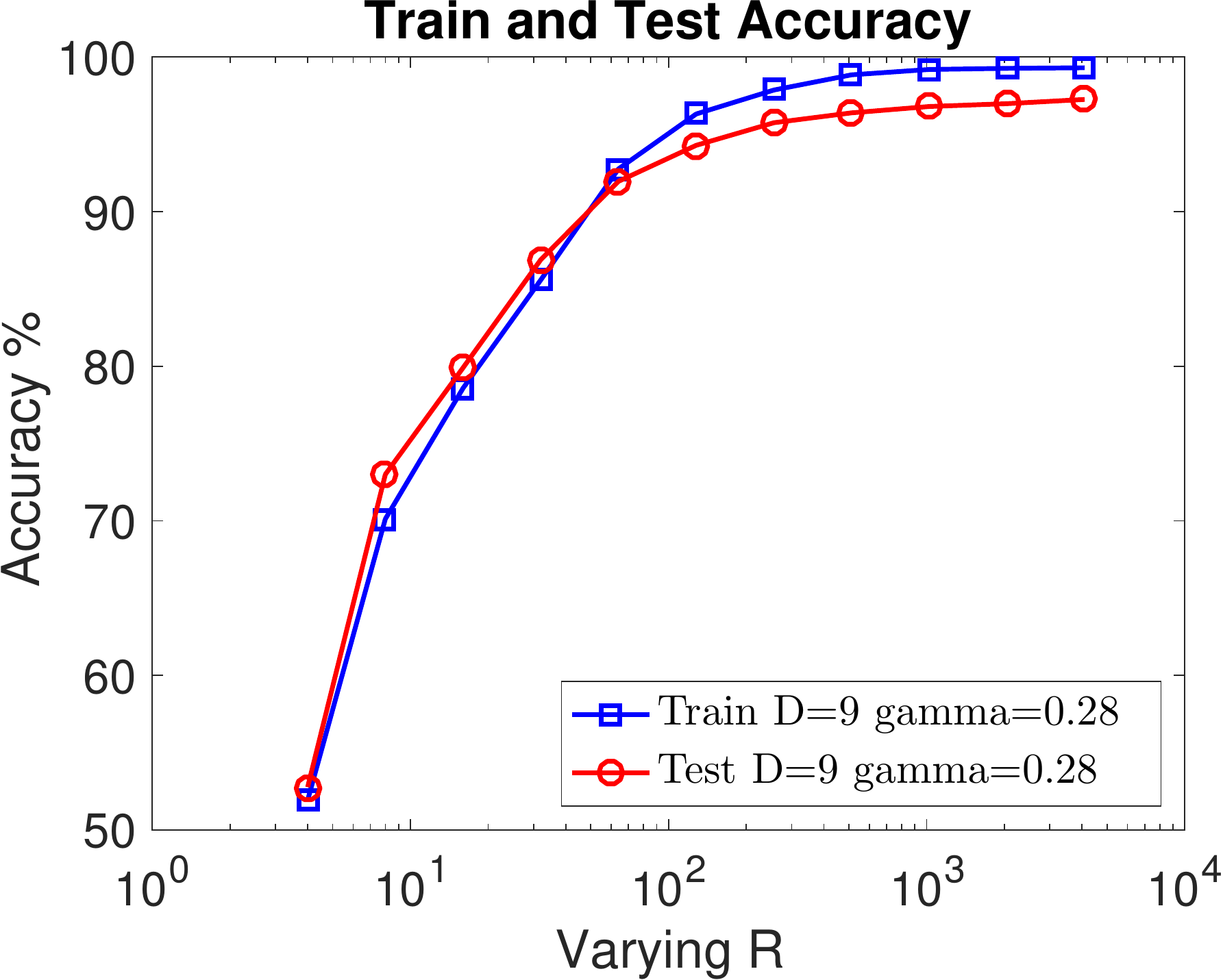}
      \caption{REUTERS}
      \label{App:fig:exptsA_varyingR_r8}
     	\end{subfigure}
		\begin{subfigure}[b]{0.23\textwidth}
      \includegraphics[width=\textwidth]{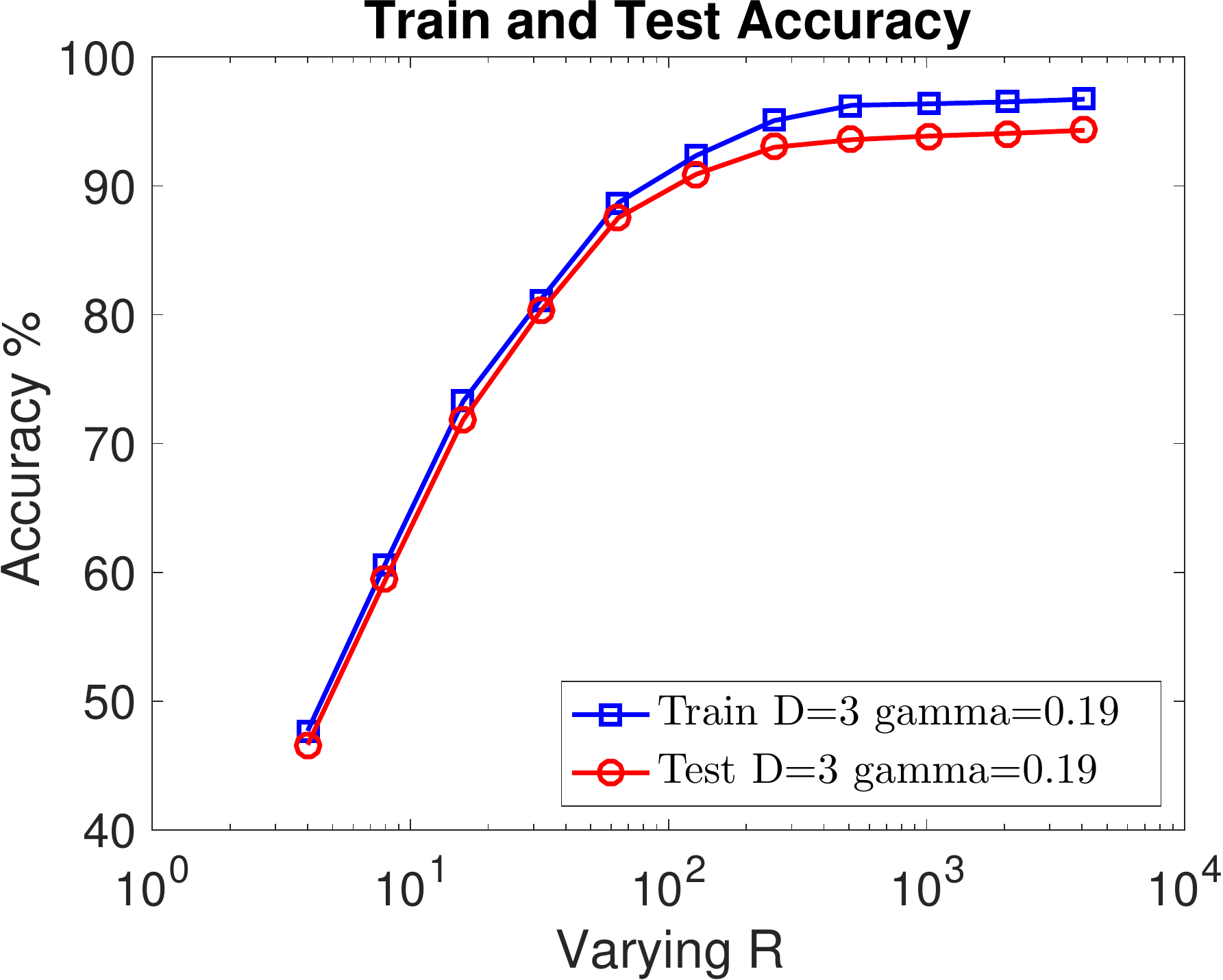}
      \caption{AMAZON}
      \label{App:fig:exptsA_varyingR_amazon}
      \end{subfigure}
		\begin{subfigure}[b]{0.23\textwidth}
      \includegraphics[width=\textwidth]{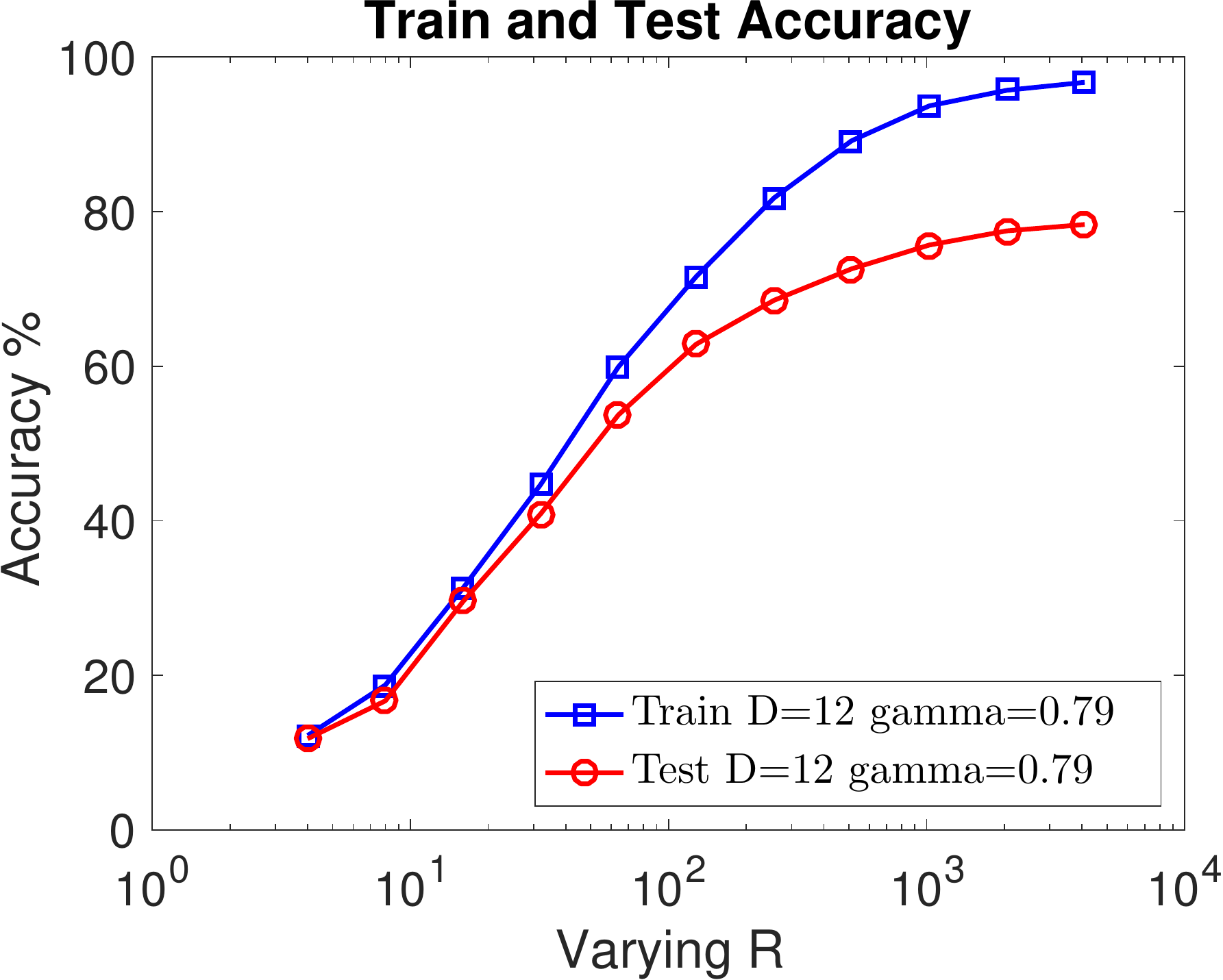}
      \caption{20NEWS}
      \label{App:fig:exptsA_varyingR_20ng2_500}
      \end{subfigure}
\caption{Train (Blue) and test (Red) accuracy when varying $R$ with fixed $D$.}
\label{App:fig:exptsA_varyingR}
\end{figure*}

\begin{figure*}[htb]
\centering
  		\begin{subfigure}[b]{0.23\textwidth}
      \includegraphics[width=\textwidth]{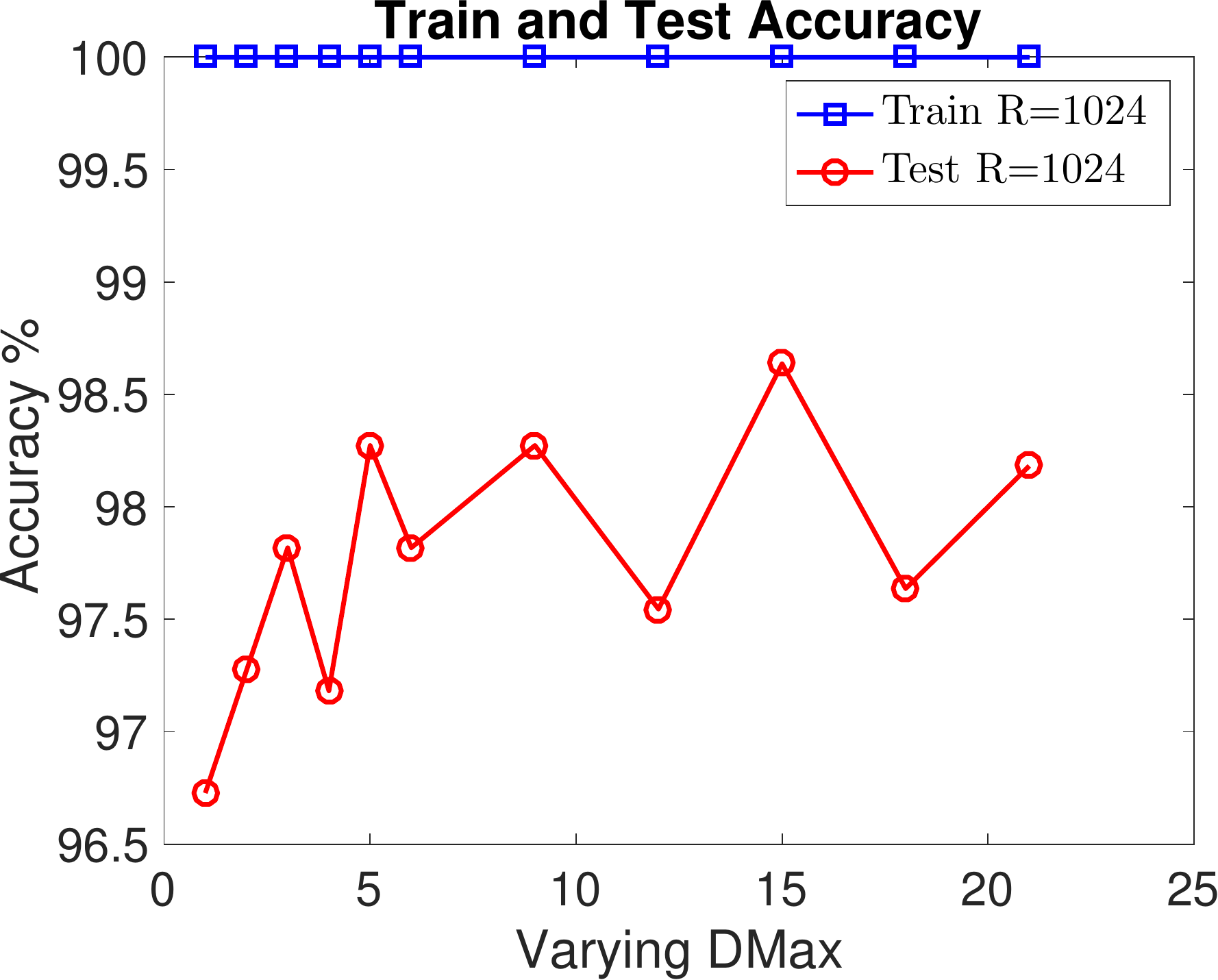}
      \caption{BBCSPORT}
      \label{App:fig:exptsA_varyingD_bbcsport}
     	\end{subfigure}
		\begin{subfigure}[b]{0.23\textwidth}
      \includegraphics[width=\textwidth]{Graphs/wmdk_varyingD/twitter_Accu_VaryingD_CV_R1024_dataSplit1-eps-converted-to.pdf}
      \caption{TWITTER}
      \label{App:fig:exptsA_varyingD_twitter}
      \end{subfigure}
		\begin{subfigure}[b]{0.23\textwidth}
      \includegraphics[width=\textwidth]{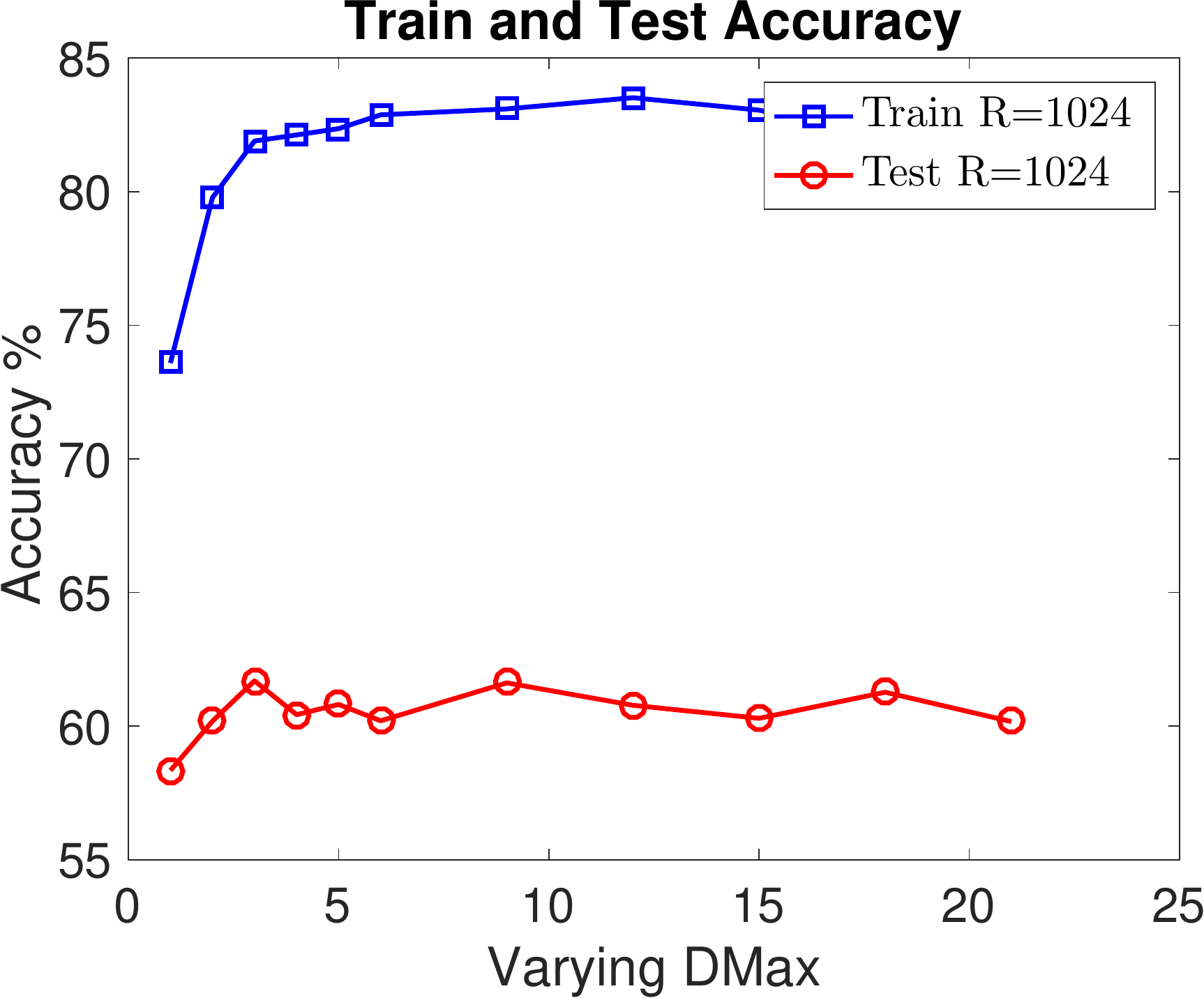}
      \caption{RECIPE}
      \label{App:fig:exptsA_varyingD_recipe2}
      \end{subfigure}
		\begin{subfigure}[b]{0.23\textwidth}
      \includegraphics[width=\textwidth]{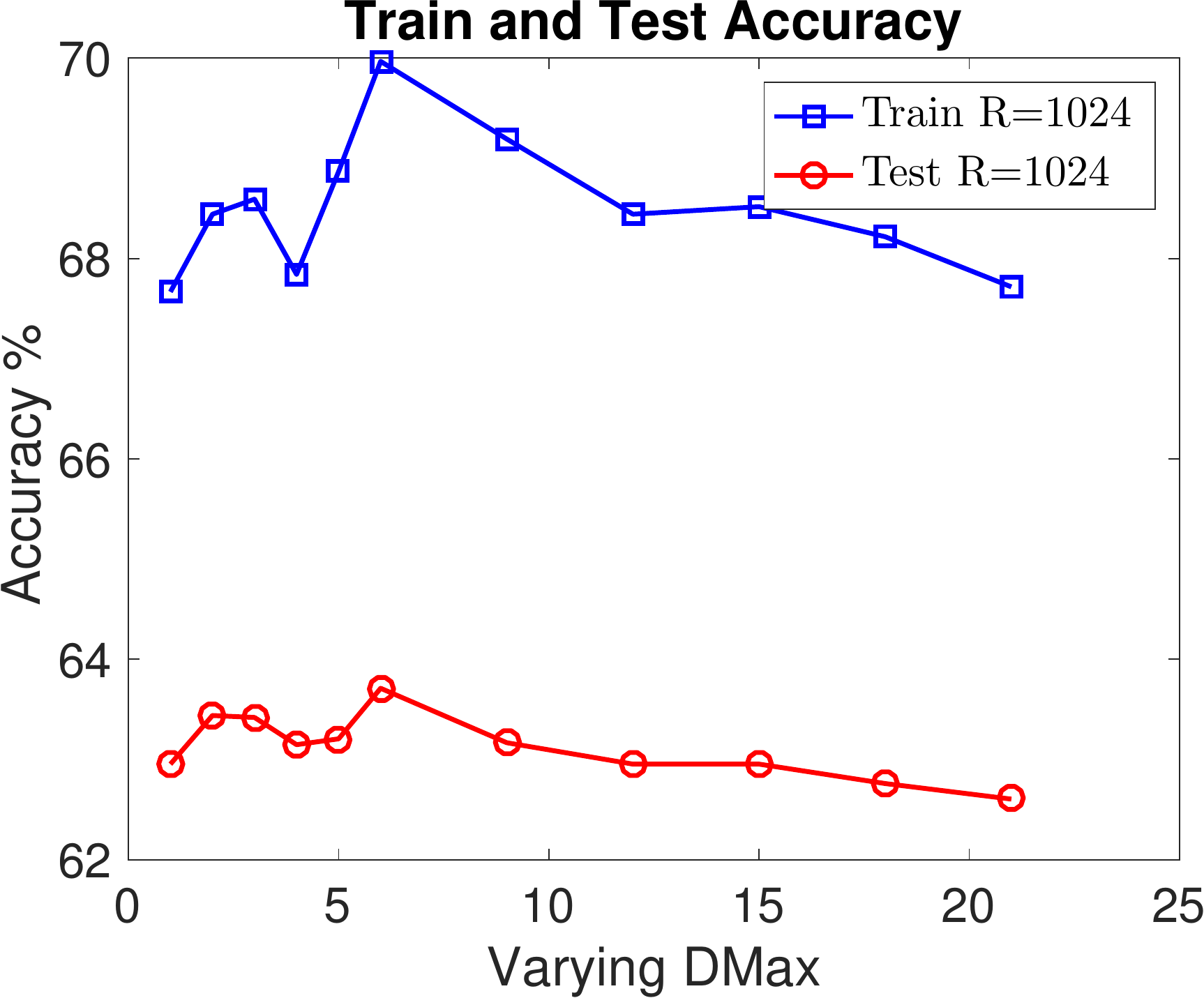}
      \caption{OHSUMED}
      \label{App:fig:exptsA_varyingD_ohsumed}
      \end{subfigure}
      \begin{subfigure}[b]{0.23\textwidth}
      \includegraphics[width=\textwidth]{Graphs/wmdk_varyingD/classic_Accu_VaryingD_CV_R1024_dataSplit1-eps-converted-to.pdf}
      \caption{CLASSIC}
      \label{App:fig:exptsA_varyingD_classic}
      \end{subfigure}
      \begin{subfigure}[b]{0.23\textwidth}
      \includegraphics[width=\textwidth]{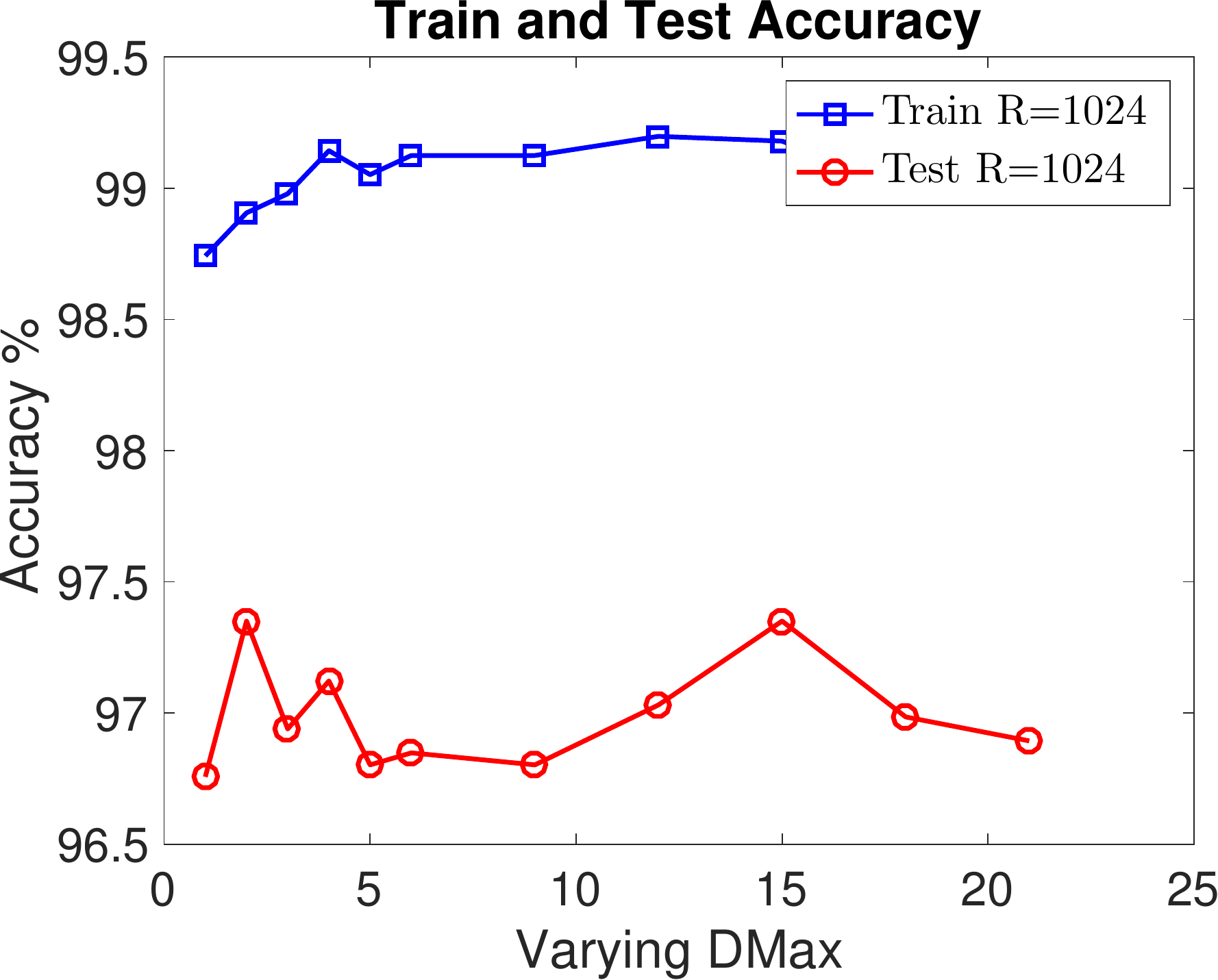}
      \caption{REUTERS}
      \label{App:fig:exptsA_varyingD_r8}
     	\end{subfigure}
		\begin{subfigure}[b]{0.23\textwidth}
      \includegraphics[width=\textwidth]{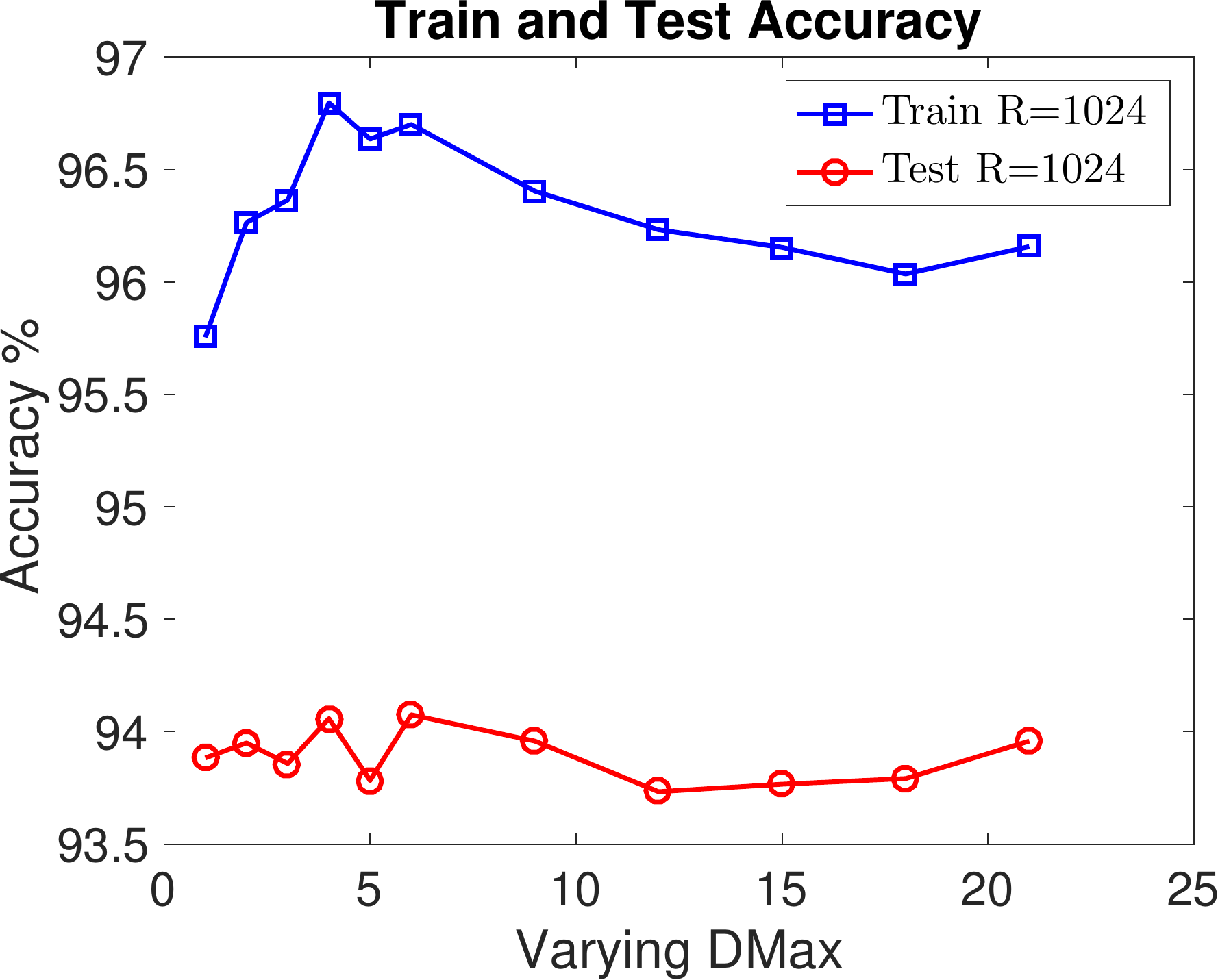}
      \caption{AMAZON}
      \label{App:fig:exptsA_varyingD_amazon}
      \end{subfigure}
		\begin{subfigure}[b]{0.23\textwidth}
      \includegraphics[width=\textwidth]{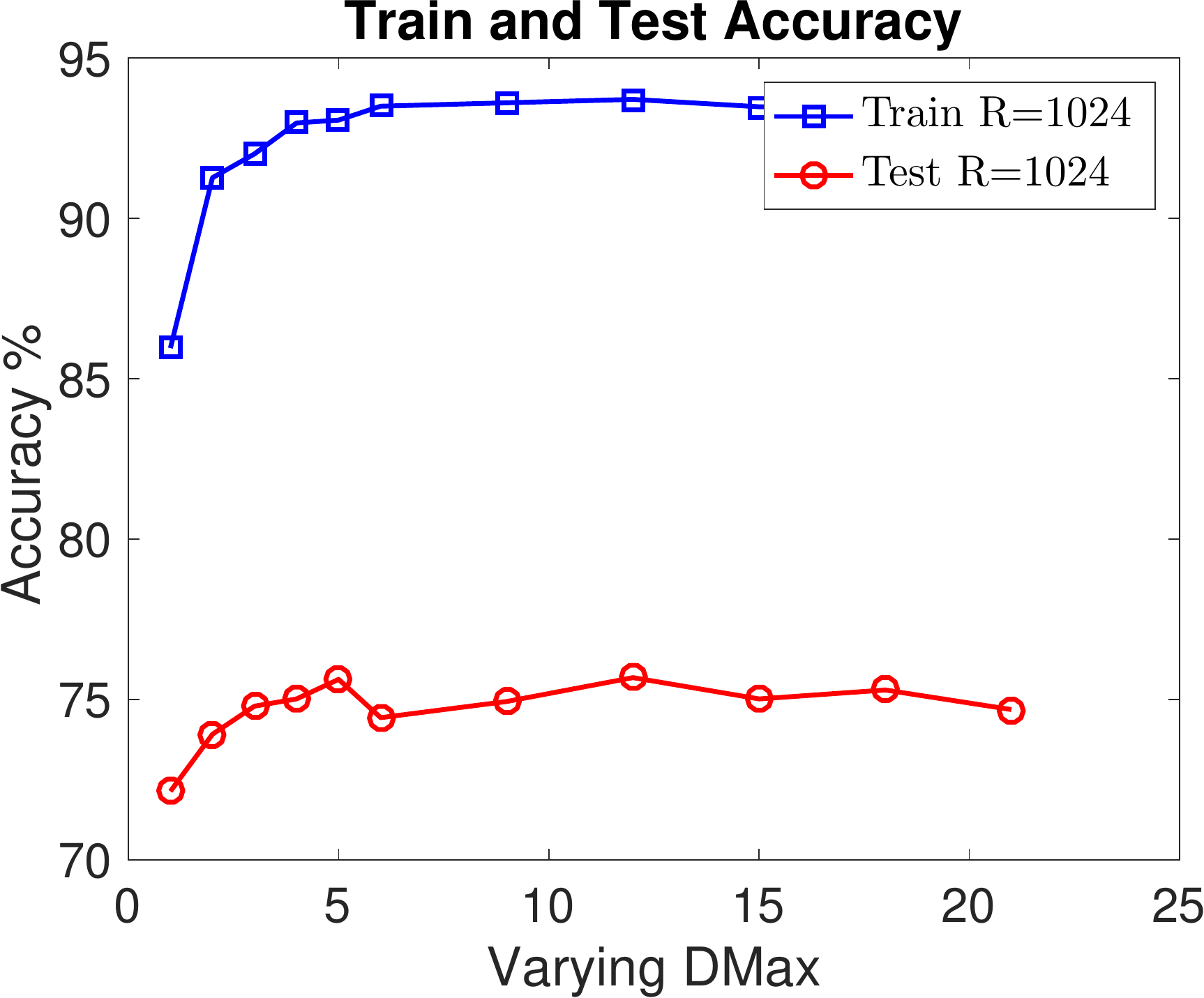}
      \caption{20NEWS}
      \label{App:fig:exptsA_varyingD_20ng2_500}
      \end{subfigure}
\caption{Train (Blue) and test (Red) accuracy when varying $D$ with fixed $R$.}
\label{App:fig:exptsA_varyingD}
\end{figure*}

\subsection{Experimental settings and parameters for WME}
\label{App:Experimental settings and parameters for WME}

\textbf{Setup.} We choose 9 different document corpora where 8 of them are overlapped with datasets in \cite{kusner2015word,huang2016supervised}. A complete data summary is in Table \ref{tb:info of datasets}. These datasets come from various applications, including news categorization, sentiment analysis, product identification, and have various number of classes, varying number of documents, and a wide range of document lengths.
Our code is implemented in Matlab and we use the C Mex function for computationally expensive components of Word Mover's Distance \footnote{We adopt Rubner's C code from \url{http://ai.stanford.edu/~rubner/emd/default.htm}. } \cite{rubner2000earth} and the freely available Word2Vec word embedding \footnote{We use word2vec code from \url{https://code.google.com/archive/p/word2vec/}. } which has pre-trained embeddings for 3 millon words/phrases (from Google News) \cite{mikolov2013efficient}. 
All computations were carried out on a DELL dual socket system with Intel Xeon processors 272 at 2.93GHz for a total of 16 cores and 250 GB of memory, running the SUSE Linux operating system. To accelerate the computation of WMD-based methods, we use multithreading with total 12 threads for WME and KNN-WMD in all experiments.
For all experiments, we generate random document from uniform distribution with mean centered in Word2Vec embedding space since we observe the best performance with this setting. We perform 10-fold cross-validation to search for best parameters for $\gamma$ and $D_{max}$ as well as parameter $C$ for LIBLINEAR on training set for each dataset. We simply fix the $D_{min}=1$, and vary $D_{max}$ in the range of 3 to 21, $\gamma$ in the range of [1e-2 3e-2 0.10 0.14 0.19 0.28 0.39 0.56 0.79 1.0 1.12 1.58 2.23 3.16 4.46 6.30 8.91 10], and $C$ in the range of [1e-5 1e-4 1e-3 1e-2 1e-1 1 1e1 1e2 3e2 5e2 8e2 1e3 3e3 5e3 8e3 1e4 3e4 5e4 8e4 1e5 3e5 5e5 8e5 1e6 1e7 1e8] respectively in all experiments. 

We collect all document corpora from these public websites:
BBCSPORT \footnote{\url{http://mlg.ucd.ie/datasets/bbc.html }}, 
TWITTER \footnote{\url{http://www.sananalytics.com/lab/twitter-sentiment/}}, 
RECIPE \footnote{\url{https://www.kaggle.com/kaggle/recipe-ingredients-dataset}},
OHSUMED \footnote{\url{https://www.mat.unical.it/OlexSuite/Datasets/SampleDataSets-download.htm}},
CLASSIC \footnote{\url{http://www.dataminingresearch.com/index.php/2010/09/classic3-classic4-datasets/}},
REUTERS and 20NEWS \footnote{\url{http://www.cs.umb.edu/~smimarog/textmining/datasets/}},
and AMAZON \footnote{\url{https://www.cs.jhu.edu/~mdredze/datasets/sentiment/}}.

\subsection{More results about effects of $R$ and $D$ on random documents}
\label{App:More results about effects of $R$ and $D$ on random documents}

\textbf{Setup and results.} To fully study the characteristic of the WME method, we study the effect of the $R$ number of random documents and the $D$ length of random documents on the performance of various datasets in terms of training and testing accuracy. Clearly, the training and testing accuracy can converge rapidly to the exact kernels when varying $R$ from 4 to 4096, which confirms our analysis in Theory 1. When varying $D$ from 1 to 21, we can see that in most of cases $D_{max} = [3, \ 12]$ generally yields a near-peak performance except BBCSPORT.

\subsection{More results on Comparisons against distance-based methods}
\label{App:More results on Comparisons against distance-based methods}

\begin{table*}[h]
\centering
\scriptsize
\caption{Testing accuracy comparing WME against KNN-based methods}
\vspace{-0mm}
\label{tb:comp_knn}
\newcommand{\Bd}[1]{\textbf{#1}}
\begin{center}
    \begin{tabular}{ c c c c c c c c c}
    \hline
    Dataset & BOW & TF-IDF & BM25 & LSI & LDA & mSDA & KNN-WMD & WME \\ \hline 
    BBCSPORT  & 79.4 $\pm$ 1.2 & 78.5 $\pm$ 2.8 & 83.1 $\pm$ 1.5 & 95.7 $\pm$ 0.6 & 93.6 $\pm$ 0.7 & 91.6 $\pm$ 0.8 & 95.4 $\pm$ 0.7 & \Bd{98.2 $\pm$ 0.6} \\ 
    TWITTER	 & 56.4 $\pm$ 0.4 & 66.8 $\pm$ 0.9 & 57.3 $\pm$ 7.8 & 68.3 $\pm$ 0.7 & 66.2 $\pm$ 0.7 & 67.7 $\pm$ 0.7 & 71.3 $\pm$ 0.6 & \Bd{74.5 $\pm$ 0.5} \\ 
    RECIPE	& 40.7 $\pm$ 1.0 & 46.4 $\pm$ 1.0 & 46.4 $\pm$ 1.9 & 54.6 $\pm$ 0.5 & 48.7 $\pm$ 0.6 & 52 $\pm$ 1.4 & 57.4 $\pm$ 0.3 & \Bd{61.8 $\pm$ 0.8} \\ 
    OHSUMED  & 38.9 & 37.3 & 33.8 & 55.8 & 49.0 & 50.7 & 55.5 & \Bd{64.5} \\ 
    CLASSIC  & 64.0 $\pm$ 0.5 & 65.0 $\pm$ 1.8 & 59.4 $\pm$ 2.7 & 93.3 $\pm$ 0.4 & 95.0 $\pm$ 0.3 & 93.1 $\pm$ 0.4 & \Bd{97.2 $\pm$ 0.1} & 97.1 $\pm$ 0.4 \\ 
    REUTERS  & 86.1 & 70.9 & 67.2 & 93.7 & 93.1 & 91.9 & 96.5 & \Bd{97.2} \\ 
    AMAZON  & 71.5 $\pm$ 0.5 & 58.5 $\pm$ 1.2 & 41.2 $\pm$ 2.6 & 90.7 $\pm$ 0.4 & 88.2 $\pm$ 0.6 & 82.9 $\pm$ 0.4 & 92.6 $\pm$ 0.3 & \Bd{94.3 $\pm$ 0.4} \\ 
    20NEWS  & 42.2 & 45.6 & 44.1 & 71.1 & 68.5 & 60.5 & 73.2 & \Bd{78.3}  \\ \hline
    \end{tabular}
\end{center}
\vspace{0mm}
\end{table*}

\begin{table*}[h]
\centering
\scriptsize
\caption{Testing accuracy of WME against Word2Vec and Doc2Vec-based methods.}
\label{App:tb:comp_word2vec}
\newcommand{\Bd}[1]{\textbf{#1}}
\begin{center}
    \begin{tabular}{ c c c c c c c c}
    \hline
    Dataset & Word2Vec+nbow & Word2Vec+tf-idf & PV-DBOW & PV-DM & Doc2VecC(Train) & Doc2VecC & WME \\ \hline 
    BBCSPORT  & 97.3 $\pm$ 0.9 & 96.9 $\pm$ 1.1 & 97.2 $\pm$ 0.7 & 97.9 $\pm$ 1.3 & 89.2 $\pm$ 1.4 & 90.5 $\pm$ 1.7 & \Bd{98.2 $\pm$ 0.6} \\ 
    TWITTER  & 72.0 $\pm$ 1.5 & 71.9 $\pm$ 0.7 & 67.8 $\pm$ 0.4 & 67.3 $\pm$ 0.3 & 69.8 $\pm$ 0.9 & 71.0 $\pm$ 0.4 & \Bd{74.5 $\pm$ 0.5} \\ 
    OHSUMED & 63.0 & 60.6 & 55.9 & 59.8 & 59.6 & 63.4 & \Bd{64.5} \\ 
    CLASSIC & 95.2 $\pm$ 0.4 & 93.9$\pm$ 0.4 & 97.0 $\pm$ 0.3 & 96.5 $\pm$ 0.7 & 96.2 $\pm$ 0.5 & 96.6 $\pm$ 0.4  & \Bd{97.1 $\pm$ 0.4} \\ 
    REUTERS & 96.9 & 95.9 & 96.3 & 94.9 & 96.0 & 96.5 & \Bd{97.2} \\ 
    AMAZON & 94.0 $\pm$ 0.5 & 92.2 $\pm$ 0.4 & 89.2 $\pm$ 0.3 & 88.6 $\pm$ 0.4 & 89.5 $\pm$ 0.4 & 91.2 $\pm$ 0.5 & \Bd{94.3 $\pm$ 0.4} \\ 
    20NEWS & 71.7  & 70.2  & 71.0 & 74.0 & 72.9 & 78.2 & \Bd{78.3} \\ 
    RECIPE\_L  & 74.9 $\pm$ 0.5 & 73.1 $\pm$ 0.6 & 73.1 $\pm$ 0.5 & 71.1 $\pm$ 0.4 & 75.6 $\pm$ 0.4  & 76.1 $\pm$ 0.4  & \Bd{79.2 $\pm$ 0.3} \\ \hline
    \end{tabular}
\end{center}
\end{table*}

\begin{table*}[h]
\centering
\scriptsize
\caption{Testing accuracy of WME against other document representations on Imdb dataset (50K). Results are collected from \cite{Chen2017efficient} and \cite{arora2017simple}.}
\vspace{-0mm}
\label{tb:comp_imdb}
\newcommand{\Bd}[1]{\textbf{#1}}
\begin{center}
    \begin{tabular}{ c c c c c c c c c}
    \hline
    Dataset & RNN\_LM & SIF(GloVe) & Word2Vec+AVG & Word2Vec+IDF & PV-DBOW & ST & Doc2VecC & WME \\ \hline 
    Imdb & 86.4 & 85.0 & 87.3 & 88.1 & 87.9 & 82.6 & 88.3 & \Bd{88.5} \\ \hline
    \end{tabular}
\end{center}
\vspace{0mm}
\end{table*}

\begin{table*}[htb]
\centering
\scriptsize
\caption{Pearson's scores of WME against other unsupervised and supervised methods on 22 textual similarity tasks. Results are collected from \cite{arora2017simple} except our approach.}
\vspace{0mm}
\label{tb:comp_textual_similarity_full}
\newcommand{\Bd}[1]{\textbf{#1}}
\begin{center}
    \begin{tabular}{ c | cccccc | ccccc | cc}
    \hline
    \multicolumn{1}{c}{Approaches}  
    & \multicolumn{6}{c}{Supervised} 
    & \multicolumn{5}{c}{Unsupervised}
    & \multicolumn{2}{c}{Semi-supervised}\\ \hline 
    \multicolumn{1}{c}{WordEmbeddings}  
    & \multicolumn{6}{c}{PSL} 
    & \multicolumn{5}{c}{GloVe}
    & \multicolumn{2}{c}{PSL}\\ \hline 
    Tasks & PP & Dan & RNN & iRNN & LSTM(no) & LSTM(o.g.) & ST & nbow & tf-idf & SIF & WME & SIF & WME \\ \hline 
    MSRpar & 42.6 & 40.3 & 18.6 & 43.4 & 16.1 & 9.3 & 16.8 & 47.7 & \Bd{50.3} & 35.6 & 45.3 & 43.3 & 49.3 \\
    MSRvid & 74.5 & 70.0 & 66.5 & 73.4 & 71.3 & 71.3 & 41.7 & 63.9 & 77.9 & 83.8 & 75.9 & \Bd{84.1} & 76.8 \\
    SMT-eur & 47.3 & 43.8 & 40.9 & 47.1 & 41.8 & 44.3 & 35.2 & 46.0 & 54.7 & 49.9 & \Bd{57.7} & 44.8 & 55.6\\
    OnWN & 70.6 & 65.9 & 63.1 & 70.1 & 65.2 & 56.4 & 29.7 & 55.1 & 64.7 & 66.2 & 67.8 & \Bd{71.8} & 69.9\\
    SMT-news & 58.4 & 60.0 & 51.3 & 58.1 & 60.8 & 51.0 & 30.8 & 49.6 & 45.7 & 45.6 & 56.1 & 53.6 & \Bd{62.5} \\
    STS'12 & 58.7 & 56.0 & 48.1 & 58.4 & 51.0 & 46.4 & 30.8 & 52.5 & 58.7 & 56.2 & 60.6 & 59.5 & \Bd{62.8}\\ \hline
    headline & 72.4 & 71.2 & 59.5 & 72.8 & 57.4 & 48.5 & 34.6 & 63.8 & 69.2 & 69.2 & 70.5 & 74.1 & \Bd{74.2}\\
    OnWN & 67.7 & 64.1 & 54.6 & 69.4 & 68.5 & 50.4 & 10.0 & 49.0 & 72.9 & \Bd{82.8} & 80.1 & 82.0 & 81.9\\
    FNWN & 43.9 & 43.1 & 30.9 & 45.3 & 24.7 & 38.4 & 30.4 & 34.2 & 36.6 & 39.4 & 33.7 & \Bd{52.4} & 32.5\\
    SMT & 39.2 & 38.3 & 33.8 & \Bd{39.4} & 30.1 & 28.8 & 24.3 & 22.3 & 29.6 & 37.9 & 33.7 & 38.5 & 36.7\\
    STS'13 & 55.8 & 54.2 & 44.7 & 56.7 & 45.2 & 41.5 & 24.8 & 42.3 & 52.1 & 56.6 & 54.5 & \Bd{61.8} & 56.3 \\ \hline
    deft forum & 48.7 & 49.0 & 41.5 & 49.0 & 44.2 & 46.1 & 12.9 & 27.1 & 37.5 & 41.2 & 41.2 & \Bd{51.4} & 45.4\\
    deft news & \Bd{73.1} & 71.7 & 53.7 & 72.4 & 52.8 & 39.1 & 23.5 & 68.0 & 68.7 & 69.4 & 66.7 & 72.6 & 69.2 \\
    headline & 69.7 & 69.2 & 57.5 & 70.2 & 57.5 & 50.9 & 37.8 & 59.5 & 63.7 & 64.7 & 65.6 & 70.1 & \Bd{71.6} \\
    images & 78.5 & 76.9 & 67.6 & 78.2 & 68.5 & 62.9& 51.2 & 61.0 & 72.5 & 82.6 & 69.2 & \Bd{84.8} & 71.4 \\
    OnWN & 78.8 & 75.7 & 67.7 & 78.8 & 76.9& 61.7 & 23.3 & 58.4 & 75.2 & 82.8 & 81.1 & \Bd{84.5} & 82.3 \\
    tweet news & 76.4 & 74.2 & 58.0 & 76.9 & 58.7 & 48.2 & 39.9 & 51.2 & 65.1 & 70.1 & 68.9 & \Bd{77.5} & 68.3 \\
    STS'14 & 70.9 & 69.5 & 57.7 & 70.9 & 59.8 & 51.5 & 31.4 & 54.2 & 63.8 & 68.5 & 65.5 & \Bd{73.5} & 68.0 \\ \hline
    answers-forum & 68.3 & 62.6 & 32.8 & 67.4 & 51.9 & 50.7 & 36.1 & 30.5 & 45.6 & 63.9 & 56.4 & \Bd{70.1} & 57.8 \\
    answers-student & \Bd{78.2} & 78.1 & 64.7 & 78.1 & 71.5 & 55.7 & 33.0 & 63.0 & 63.9 & 70.4 & 63.1 & 75.9 & 66.2 \\ 
    belief & \Bd{76.2} & 72.0 & 51.9 & 75.9 & 61.7 & 52.6 & 24.6 & 40.5 & 49.5 & 71.8 & 50.6 & 75.3 & 51.6 \\
    headline & 74.8 & 73.5 & 65.3 & 75.1 & 64.0 & 56.6 & 43.6 & 61.8 & 70.9 & 70.7 & 70.8 & 75.9 & \Bd{76.1} \\
    images & 81.4 & 77.5 & 71.4 & 81.1 & 70.4 & 64.2 & 17.7 & 67.5 & 72.9 & 81.5 & 67.9 & \Bd{84.1} & 69.3 \\
    STS'15 & 75.8 & 72.7 & 57.2 & 75.6 & 63.9 & 56.0 & 31.0 & 52.7 & 60.6 & 71.7 & 61.8 & \Bd{76.3} & 64.2 \\ \hline
    SICK'14 & 71.6 & 70.7 & 61.2 & 71.2 & 63.9 & 59.0 & 49.8 & 65.9 & 69.4 & 72.2 & 68.0 & \Bd{72.9} & 68.1 \\ \hline
    Twitter'15 & 52.9 & \Bd{53.7} & 45.1 & 52.9 & 47.6 & 36.1 & 24.7 & 30.3 & 33.8 & 48.0 & 41.6 & 49.0 & 47.4 \\ \hline
    \end{tabular}
\end{center}
\vspace{0mm}
\end{table*}

\textbf{Setup.} We preprocess all datasets by removing all words in the SMART stop word list \cite{buckley1995automatic}. For 20NEWS, we remove the words appearing less than 5 times. For LDA, we use the Matlab Topic Modeling Toolbox \cite{griffiths2007probabilistic} and use sample code that first run 100 burn-in iterations and then run the chain for additional 1000 iterations. 
For mSDA, we use high-dimensional function mSDAhd where the parameter dd is set as 0.2 times BOW Dimension. For all datasets, a 10-fold cross validation on training set is performed to get the optimal $K$ for KNN classifier, where $K$ is searched in the range of $[1, 21]$. 

\textbf{Baselines.} We compare against 7 document representation or distance methods: 1) \emph{bag-of-words} (BOW) \cite{salton1988term}; 2) \emph{term frequency-inverse document frequency} (TF-IDF) \cite{robertson1994some}; 3) \emph{Okapi BM25} \cite{robertson1995okapi}: first TF-IDF variant ranking function used in search engines; 4) \emph{Latent Semantic Indexing} (LSI) \cite{deerwester1990indexing}: factorize BOW into their leading singular components subspace using SVD \cite{wu2015preconditioned,wu2017primme_svds}; 5) \emph{Latent Dirichlet Allocation} (LDA) \cite{blei2003latent}: a generative probability method to model mixtures of word "topics" in documents. LDA is trained \emph{transductively} on both train and test; 6) \emph{Marginalized Stacked Denoising Autoencoders} (mSDA) \cite{chen2012marginalized}: a fast method for training denoising autoencoder that achieved state-of-the-art performance on sentiment analysis tasks \cite{glorot2011domain}; 7) \emph{WMD}: a state-of-the-art document distance discussed in Section \ref{sec:Word2Vec and WMD}.  

\textbf{Results.} Table \ref{tb:comp_knn} clearly demonstrates the superior performance of our method WME compared to other KNN-based methods in terms of testing accuracy. Indeed, BOW and TF-IDF performs poorly compared to other methods which may be the result of frequent near-orthogonality of their high-dimensional sparse feature representation in KNN classifier. KNN-WMD achieves noticeably better testing accuracy than LSI, LDA and mSDA since WMD takes into account the word alignments and leverages the power of Word2Vec. Remarkably, our proposed method WME achieves much higher accuracy compared to other methods including KNN-WMD on all datasets except one (CLASSIC). The substantially improved accuracy of WME suggests that a truly p.d. kernel implicitly admits expressive feature representation of documents learned from the Word2Vec embedding space in which the alignments between words are considered by using WMD.

\subsection{More results on comparisons against Word2Vec and Doc2Vec-based document representations}
\label{App:More Results on Comparisons against Word2Vec and Doc2Vec-based document representations}

\textbf{Setup and results.} For \emph{PV-DBOW}, \emph{PV-DM}, and \emph{Doc2VecC}, we set the word and document vector dimension $d=300$ to match the pre-trained word embeddings we used for WME and other Word2Vec-based methods in order to make a fair comparison. For other parameters, we use recommended parameters in the papers but we search for the best parameter $C$ in LIBLINEAR for these methods. Additionally, we also train \emph{Doc2VecC} with different corruption rate in the range of [0.1 0.3 0.5 0.7 0.9]. Following \cite{Chen2017efficient}, these methods are trained transductively on both training and testing set. For \emph{Doc2VecC(Train)}, we train the model only on training set in order to show the effect of the transductive training on the testing accuracy. As shown in Table \ref{App:tb:comp_word2vec}, \emph{Doc2VecC} clearly outperforms \emph{Doc2VecC(Train)}, sometimes having a significant performance boost on some datasets (OHSUMED and 20NEWS).

We further conduct experiments on Imdb dataset using our method. We use only training data to select hyper-parameters. For a more fair comparison, we only report the results of other methods that use all data excluding test. Table \ref{tb:comp_imdb} shows that WME can achieve slightly better accuracy than other state-of-the-art document representation methods. This collaborates the importance to make full use of both word alignments and high-quality pre-trained word embeddings.

\subsection{More results on comparisons for textual similarity tasks}
\label{App:More results on comparisons for textual similarity tasks}

\textbf{Setup and results.} To obtain the hyper-parameters in our method, we use the corresponding training data or the similar tasks from previous years. Note that the tasks with same names but in different years are different ones. As we can see in Table \ref{tb:comp_textual_similarity_full}, WME can achieve better performance on tasks of STS'12 and perform fairly well on other tasks. Among the unsupervised methods and some supervised methods except \emph{PP}, \emph{Dan}, and \emph{iRNN}, WME is almost always to be one of the best methods. 

\end{document}